\documentclass[10pt]{article} 
\usepackage[preprint]{tmlr}

\usepackage{amsmath,amsfonts,bm}

\def\eqref#1{equation~\ref{#1}}

\def\1{\bm{1}}

\DeclareMathAlphabet{\mathsfit}{\encodingdefault}{\sfdefault}{m}{sl}
\SetMathAlphabet{\mathsfit}{bold}{\encodingdefault}{\sfdefault}{bx}{n}

\usepackage{graphicx}
\usepackage{amsmath}
\usepackage{amssymb}
\usepackage{booktabs}
\usepackage{comment}
\usepackage{makecell}
\usepackage{amsfonts}
\usepackage{float}
\usepackage{multirow}
\usepackage{subcaption}
\usepackage{lipsum}
\usepackage{color, colortbl}

\usepackage{hyperref}
\usepackage{url}

\title{DeblurDiNAT: A Compact Model with Exceptional \\Generalization and Visual Fidelity on Unseen Domains}

\author{\name Hanzhou Liu \email hanzhou1996@tamu.edu \\
      Texas A\&M University
      \AND
      \name Binghan Li \email lbh1994usa@tamu.edu \\
      Texas A\&M University
      \AND
      \name Chengkai Liu \email liuchengkai@tamu.edu\\
      Texas A\&M University
      \AND
      \name Mi Lu \email mlu@ece.tamu.edu  \\
      Texas A\&M University}

\newcommand\ith{\mathop{\mbox{$i$-}}}
\newcommand\bftab{\fontseries{b}\selectfont}
\def\Rplus{\raisebox{.4ex}{\tiny\bf +}}
\newcommand{\ul}[1]{\underline{#1}}
\newcommand{\ull}[1]{\underline{\underline{#1}}}

\definecolor{Gray}{gray}{0.9}

\def\month{MM}  
\def\year{YYYY} 
\def\openreview{\url{https://openreview.net/forum?id=XXXX}} 

\begin{document}

\maketitle
\begin{abstract}
Recent deblurring networks have effectively restored clear images from the blurred ones. However, they often struggle with generalization to unknown domains. Moreover, these models typically focus on distortion metrics such as PSNR and SSIM, neglecting the critical aspect of metrics aligned with human perception. To address these limitations, we propose DeblurDiNAT, a deblurring Transformer based on Dilated Neighborhood Attention. First, DeblurDiNAT employs an alternating dilation factor paradigm to capture both local and global blurred patterns, enhancing generalization and perceptual clarity. Second, a local cross-channel learner aids the Transformer block to understand the short-range relationships between adjacent channels. Additionally, we present a linear feed-forward network with a simple while effective design. Finally, a dual-stage feature fusion module is introduced as an alternative to the existing approach, which efficiently process multi-scale visual information across network levels. Compared to state-of-the-art models, our compact DeblurDiNAT demonstrates superior generalization capabilities and achieves remarkable performance in perceptual metrics, while maintaining a favorable model size. 
\end{abstract}

\section{Introduction}

Image deblurring aims to restore clarity from blurred images~\citep{cho2012text,yan2017image}. Although non-deep learning methods have shown effectiveness under certain conditions~\citep{chen2011effective,bahat2017non,hirsch2011fast,xu2010two}, their performance often falls short in real-world scenarios where blurred patterns exhibit various orientations and magnitudes~\citep{zhang2022deep}.


With the advancement of deep learning, convolutional neural networks (CNNs) have been widely employed to address image deblurring tasks~\citep{kupyn2018deblurgan, tao2018scale, purohit2020region, cho2021rethinking, zou2021sdwnet, ji2022xydeblur, fu2022edge}. To achieve a large receptive field, latest deep CNNs~\citep{chen2022simple,fang2023self} are scaled to large model sizes. More recently, Transformers~\citep{vaswani2017attention} have dominated computer vision tasks~\citep{dosovitskiy2020image, liu2021swin, yuan2021tokens, jiang2021transgan}. A key feature of Transformers is their self-attention (SA) mechanism, which inherently enables long-range feature learning. To help Transformer architectures capture local blurred patterns, several methods incorporating window partitioning and 2-D convolutions have been proposed~\citep{wang2022uformer, tsai2022stripformer, zamir2022restormer, kong2023efficient}. However, these studies often overlook short-range cross-channel interactions, leading to channel-wise information loss. This limitation hampers the networks' generalization capabilities and perceptual performance, both of which are crucial for effectively addressing real-world image deblurring challenges.

\begin{figure}[tb]
  \centering
  \includegraphics[width=0.98\linewidth, trim = 0.13cm 4.28cm 5.13cm 0.13cm]{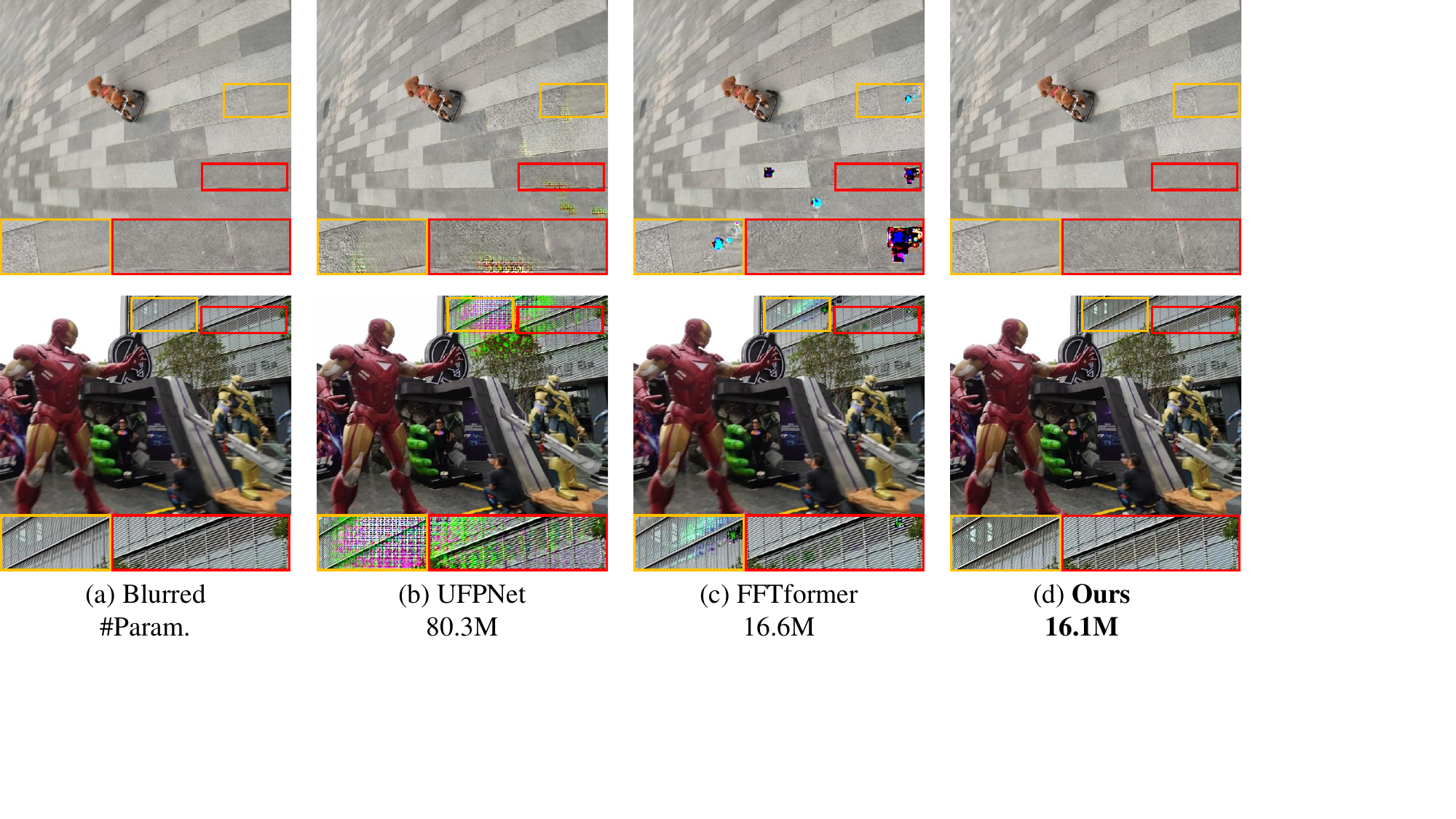}
     \caption{Deblurred results on the unseen RWBI dataset~\citep{zhang2020deblurring}, where {\bftab mode collapse} occurs in SOTA networks, UFPNet~\citep{fang2023self} and FFTformer~\citep{kong2023efficient}. In contrast, our proposed DeblurDiNAT resolves this issue while achieving comparable image quality scores with a minimal model size.}\label{fig:intro_mode_collapse}
\end{figure}

We begin by discussing the {\bftab limited generalization problem} in state-of-the-art (SOTA) deblurring approaches. A typical manifestation of this issue is {\bftab mode collapse} in image deblurring tasks, as illustrated in Figure~\ref{fig:intro_mode_collapse}. While UFPNet~\citep{fang2023self} and FFTformer~\citep{kong2023efficient} achieve SOTA performance on benchmark datasets, their generalization to real-world blurred images remains questionable. Both models produce undesirable noisy pixels regardless of the input, resulting in their inability to adapt effectively to unseen domains. Notably, such an issue has not been observed in the deblurred results of our proposed DeblurDiNAT.

\begin{figure}[tb]
  \centering
  \includegraphics[width=0.98\linewidth, trim = 0.13cm 12.5cm 7.03cm 0.13cm]{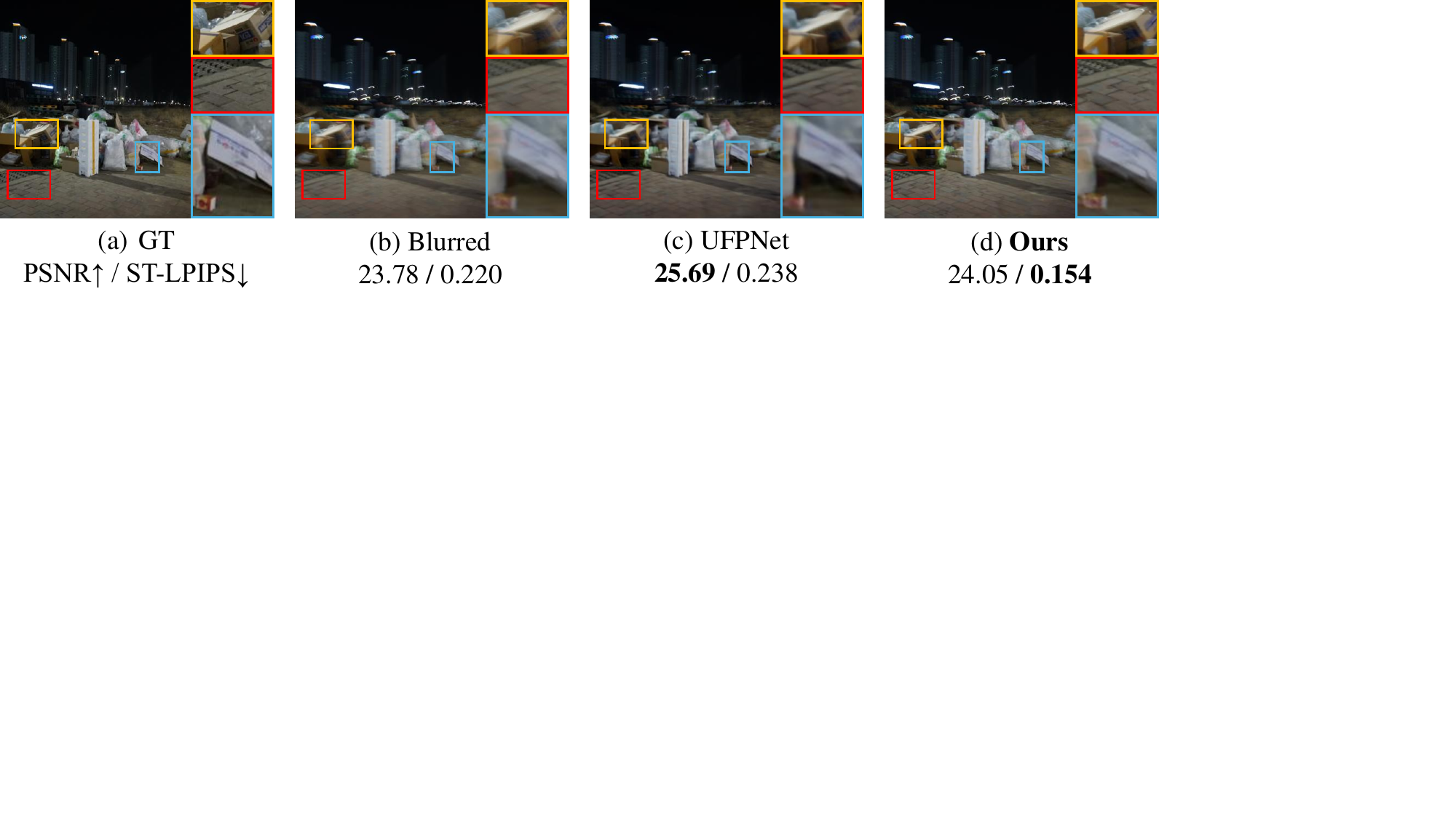}
     \caption{Which deblurred image is clearer? The above example on RealBlur-J~\citep{rim2020real} suggests that ST-LPIPS~\citep{ghildyal2022stlpips} effectively captures visual fidelity in real-world scenarios, whereas the distortion metric PSNR fails, emphasizing the importance of using perceptual metrics.}\label{fig:intro_compare_metrics}
\end{figure}

We next revisit the {\bftab limitations of distortion metrics} such as PSNR. As shown in Figure~\ref{fig:intro_compare_metrics}, our deblurred image deliver a lower PSNR but exhibits higher visual quality compared to UFPNet~\citep{fang2023self}. Distortion metrics often fail to align with perceptual quality, which is why Learned Perceptual Image Patch Similarity (LPIPS)~\citep{zhang2018unreasonable} and the enhanced alternatives like Shift-tolerant LPIPS (ST-LPIPS)~\citep{ghildyal2022stlpips} are gaining popularity for better correlating with human visual perception. In this paper, we employ ST-LPIPS since the real-world datasets contain image pairs with minor shifts. Figure~\ref{fig:intro_compare_metrics} demonstrates that our method generates significantly sharper edges and clearer textures, with LPIPS and ST-LPIPS effectively capturing these perceptual improvements, whereas PSNR fails to reflect them.

To achieve {\bftab robust generalization capabilities} and {\bftab high perceptual performance}, we develop DeblurDiNAT, a Transformer architecture for real-world image deblurring. First, DeblurDiNAT is equipped with alternating local and global Transformer blocks, which are built upon dilated neighborhood attention (DiNA)~\citep{hassani2022dilated,hassani2023neighborhood} with small and large dilation factors. This structure with alternating dilation factors provides flexible receptive fields to capture blurred patterns with varying magnitutes.

To assist Transformer architectures to better understand cross-channel dependencies, we introduce a local cross-channel learner (LCCL) that only considers each channel's nearest neighbors. LCCL draws inspiration from channel attention~\citep{hu2018squeeze,wang2020eca}. Building on this novelty, we propose a new self-attention framework, termed as channel-aware self-attention (CASA), where a parallel LCCL adaptively modulates the outputs of self-attention layers based on their shared inputs. Point-wise 2D convolutions can help capture global channel relationships, which is not the focus of this paper. 

The other basic component of Transformers is the feed-forward networks (FFNs). Inspried by the simple gate without any non-linear activation in CNN-based NAFNet~\citep{chen2022simple}, we remove the GELU function~\citep{hendrycks2016gaussian} from the gated-dconv feed-forward network (GDFN)~\citep{zamir2022restormer}, and simply it to a linear divide and multiply feed-forward network (DMFN). It is worth mentioning that the non-linearity is ensured in other units of DeblurDiNAT, such as encoders and feature fusion modules.

In addition to the above novelties, we design a novel feature fusion paradigm for DeblurDiNAT. As no activation exists in our DMFN of decoders, non-linearity is introduced before the decoder path to effectively model complex visual data. To achieve this, we propose the non-linear lightweight dual-stage feature fusion (LDFF), an efficient method for merging and propagate visual features across the networks.

Our contributions are summarized as follows:
\begin{itemize}
  \item We design a local and global blur learning strategy by alternating different dilation factors of DiNA, integrated with a local cross-channel learner to capture adjacent channel relationships. 
  \item We present a simple while effective feed-forward network and a lightweight dual-stage feature fusion method for our Transformer architecture to effectively process complex blur information.
  \item Our proposed compact network, DeblurDiNAT, exhibits exceptional generalization capabilities and delivers high perceptual quality, while preserving a lightweight model design.
\end{itemize}

\section{Related Work}


\noindent \textbf{CNNs for Image Deblurring.}
In the past decade, CNN-based methods have demonstrated significant effectiveness in image deblurring tasks~\citep{hradivs2015convolutional, svoboda2016cnn, nah2017deep, kupyn2019deblurgan, tao2018scale, gao2019dynamic, zhang2019deep, cho2021rethinking, zamir2021multi, chen2022simple, fang2023self}. MIMO-UNet~\citep{cho2021rethinking} introduces a concatenate-and-convolve strategy to combine multi-scale and same-scale features within a U-Net architecture. MPRNet~\citep{zamir2021multi} innovatively proposes a multi-stage progressive pipeline that significantly enhances deblurring performance. NAFNet~\citep{chen2022simple} presents a simple yet effective baseline for high-quality image restoration, while UFPNet~\citep{fang2023self} further improves deblurring performance by incorporating kernel estimation. To capture global blurred patterns, the latest CNN-based methods, including NAFNet and UFPNet~\citep{chen2022simple, fang2023self}, have been scaled to very large sizes to achieve a large receptive field.

\noindent \textbf{Transformers for Image Deblurring.}
More recently, Transformer networks have been studies for the image deblurring problem~\citep{wang2022uformer,tsai2022stripformer,zamir2022restormer,kong2023efficient,mao2024loformer}. Restormer~\citep{zamir2022restormer} computes cross-covariance across feature channels to model global connectivity. 
FFTformer~\citep{kong2023efficient}, a novel frequency-based Transformer, achieves state-of-the-art (SOTA) performance on the known domains. The latest LoFormer~\citep{mao2024loformer} incorporates channel-wise self-attention in the frequency domain and achieves deblurred results comparable to FFTformer. However, these Transformer-based approaches lack exploration on model generalization to unseen real-world domains, and they do not adequately quantify perceptual performance either.

\section{Method}

\begin{figure*}[tb]
  \centering
  \includegraphics[width=0.98\linewidth, trim = 0.3cm 5.1cm 0.2cm 0.3cm]{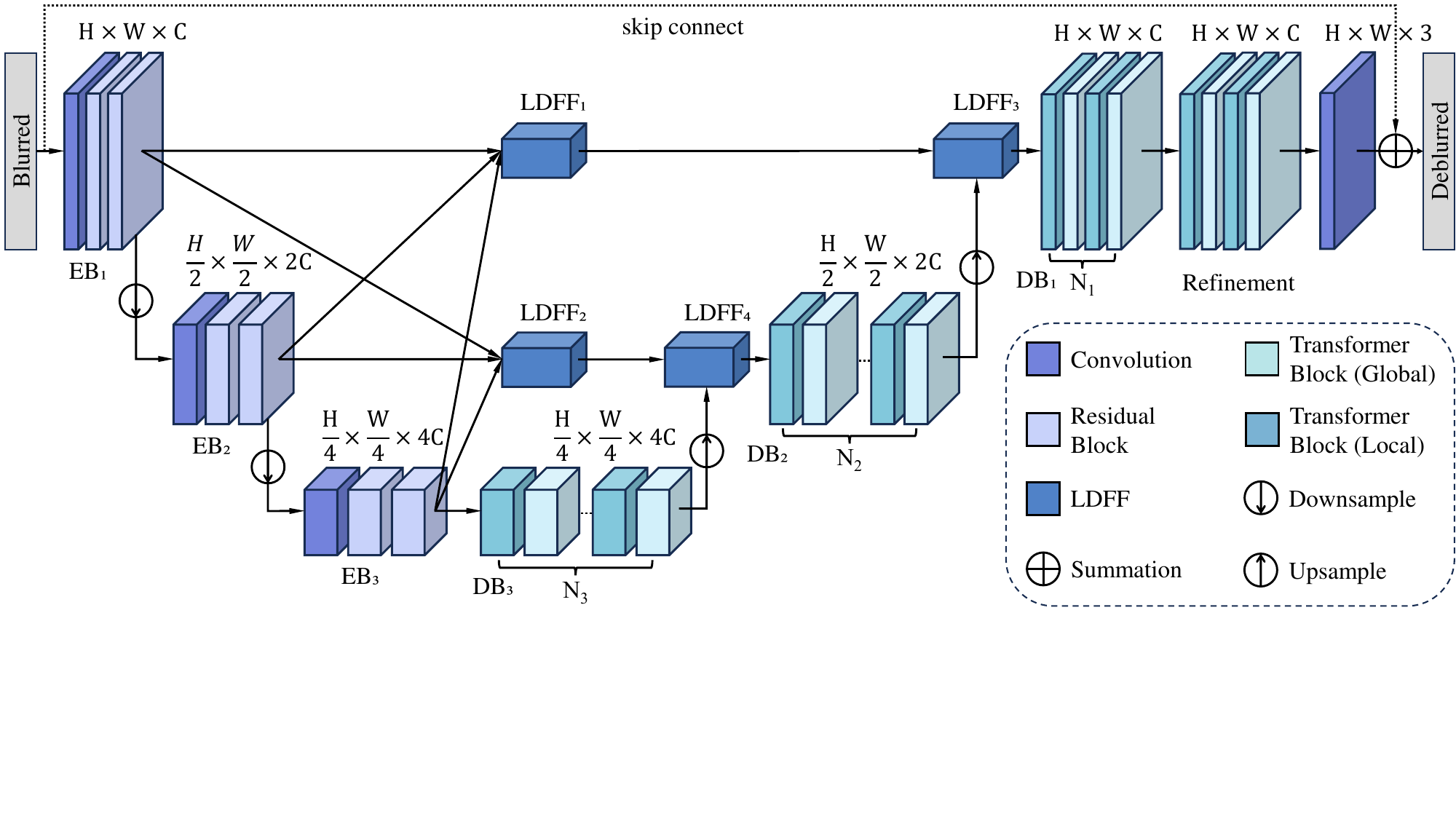}

  \caption{Architecture of DeblurDiNAT for single image deblurring. DeblurDiNAT is a hybrid encoder-decoder Transformer, where cascaded residual blocks extract multi-scale image features from input images, and the alternating local and global Transformer blocks reconstructs clean images from hierarchical features.
  }\label{fig:method_architecture}
\end{figure*}

\noindent \textbf{Preliminaries.}
We adopt Dilated Neighborhood Attention (DiNA)~\citep{hassani2022dilated, hassani2023neighborhood} for image deblurring. DiNA serves as a flexible SA mechanism for short- and long-range learning by adjusting the dilation factor
without additional complexity theoretically.
For simplicity, consider a feature map $\mathbf{X}\in\mathbb{R}^{n\times d}$,
where $n$ is the number of tokens (basic data units) and $d$ is the dimension;
the $query$ and $key$ linear projections of $\mathbf{X}$, $Q$ and $K$;
and relative positional biases between two tokens $i$ and $j$, $B(i, j)$.
Given a dilation factor $\delta$ and neighborhood size $k$,
the attention weights for the $\ith$th token is defined as:
\begin{equation}
  \begin{aligned} \label{eq_qk}
    \mathbf{A}_i^{(k, \delta)} = \begin{bmatrix}
    Q_{i}K_{\rho_1^{\delta}(i)}^{T}+B_{(i,\rho_{1}^{\delta}(i))}\\
    Q_{i}K_{\rho_2^{\delta}(i)}^{T}+B_{(i,\rho_{2}^{\delta}(i))}\\
    \vdots\\
    Q_{i}K_{\rho_k^{\delta}(i)}^{T}+B_{(i,\rho_{k}^{\delta}(i))}
  \end{bmatrix}.
  \end{aligned}
\end{equation}
The corresponding matrix $\mathbf{V}_i^{(k, \delta)}$,
whose elements are the $\ith$th token's $k$ adjacent $value$ linear projections, 
can be obtained by:
\begin{equation}
  \begin{aligned} \label{eq_v}
    \mathbf{V}_i^{(k, \delta)} = \begin{bmatrix}
      V_{\rho_1^{\delta}(i)}^T & V_{\rho_2^{\delta}(i)}^T & \cdots & V_{\rho_k^{\delta}(i)}^T
    \end{bmatrix}^T.
  \end{aligned}
\end{equation}
The output feature map of DiNA for the $\ith$th token is achieved by:
\begin{equation}
  \begin{aligned} \label{eq_dina}
    \mathrm{DiNA}_k^{\delta}(i) = \emph{softmax} \: (\frac{\mathbf{A}_i^{(k, \delta)}}{\sqrt{d_k}}) \mathbf{V}_i^{(k, \delta)}.
  \end{aligned}
\end{equation}

\definecolor{darkblue}{RGB}{46,117,182}
\definecolor{darkgreen}{RGB}{84,130,53}
\subsection{Alternating Dilation Factor Structure}\label{sec:adfs}

\begin{figure*}[tb]
  \centering
  \includegraphics[width=0.98\linewidth, trim = 0.5cm 13.2cm 4.05cm 0.62cm]{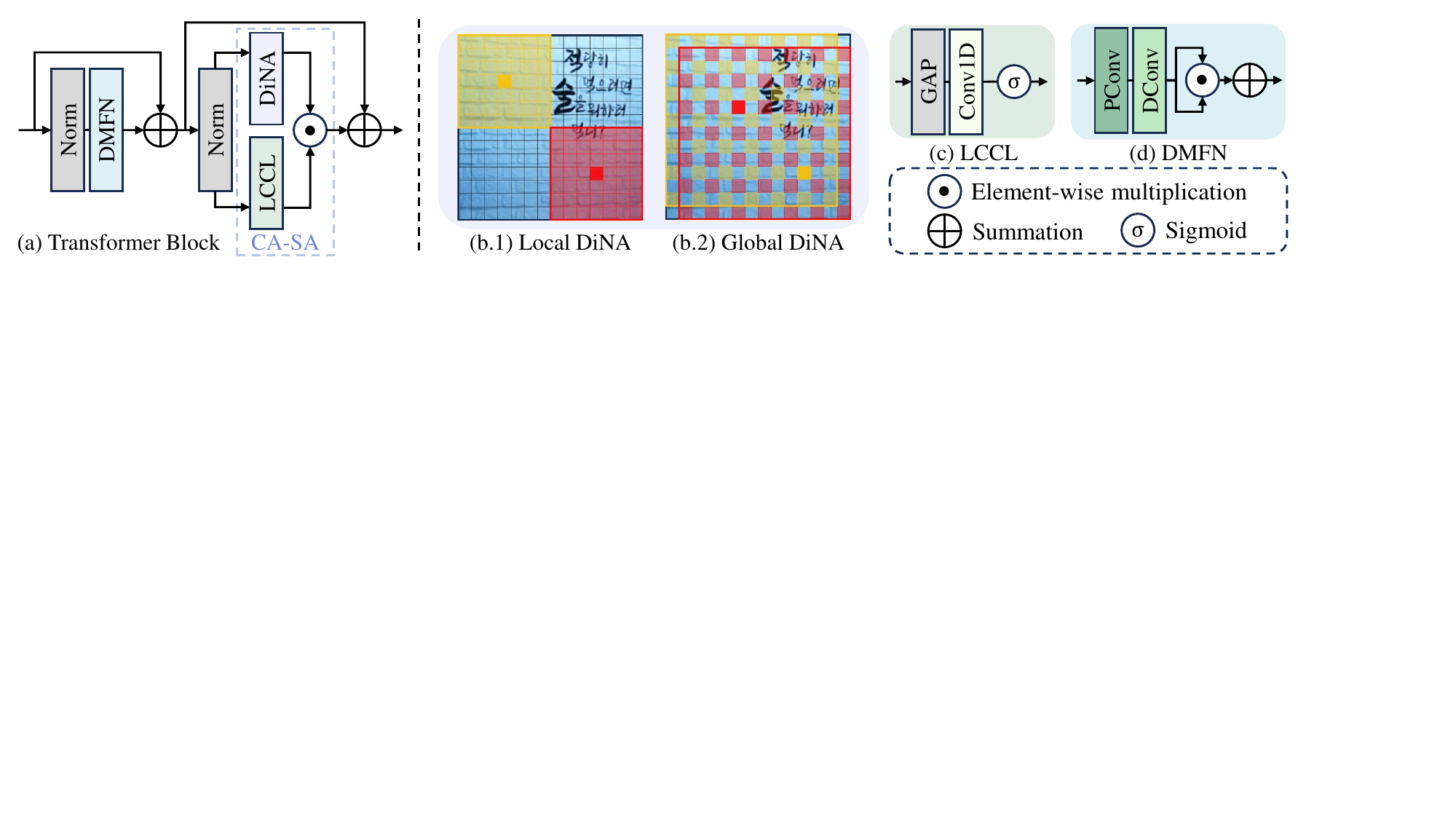}

  \caption{Structures of (a) Transformer block, (b.1) local DiNA, (b.2) global DiNA, (c) local cross-channel learner (LCCL), and (d) divide-and-multiply feed-forward network (DMFN) in DeblurDiNAT. Note: GAP represents global average pooling; PConv and Dconv refer to point-wise and depth-wise convolutions; Conv1D denotes the convolution applied to 1D sequences. Tensor transformations are omitted for simplicity. 
  }\label{fig:method_blocks}
\end{figure*}

As shown in Figure~\ref{fig:method_architecture} and~\ref{fig:method_blocks} (b.1) and (b.2), to capture both local and global blurred patterns, the decoder path of DeblurDiNAT alternates the Transformer blocks containing DiNA with a dilation factor of $\delta \in \left\{1, \lfloor\frac{n}{k}\rfloor\right\}$, where $n$ is the feature size, and $k$ is the kernel size set to a constant number $7$ in our full model. The self-attention window size is widened by incrementing the value of dilation factor, enabling the network to efficiently learn larger-area blurs.

\subsection{Channel Aware Self Attention}
Figure~\ref{fig:method_blocks} (a) shows that channel aware self-attention (CASA) contains two parallel units, DiNA and a local cross-channel learner (LCCL). As illustrated in Figure~\ref{fig:method_blocks} (c), an LCCL first transforms the normalized 2-D features into 1-D data by global average pooling (GAP); then, it applies a 1-D convolution to the intermediate features along the channel dimension; finally, a $\emph{sigmoid}$ function is adopted to compute attention scores. The outputs of the LCCL and DiNA\@ are merged by element-wise multiplications.

Given a layer normalized input tensor $\mathbf{X}\in\mathbb{R}^{\hat{H}\times\hat{W}\times\hat{C}}$.
The output of CASA is computed as,
\begin{equation}
  \begin{aligned} \label{eq_LCCL}
    \hat{\mathbf{X}} &= \mathrm{DiNA}(\mathbf{X})\odot \mathrm{LCCL}(\mathbf{X}),\\
  {\rm LCCL}(\mathbf{X}) &= f^{-1}(\mathrm{Conv}_{1d}(f(\mathrm{GAP}_{2d}^{1\times1}(\mathbf{X})))),
  \end{aligned}
\end{equation}
where $\odot$ denotes element-wise multiplication; $f$ is a tensor manipulation function which squeezes and transposes a $C\times 1\times 1$ matrix, resulting in a $1\times C$ matrix; $\mathrm{Conv}_{1d}$ denotes a 1D convolution with a kernel size of $3$; ${\rm GAP}_{2d}^{1\times1}$ indicates global average pooling, outputting a tensor of size $1\times1$.


\subsection{Divide and Multiply Feed-Forward Network}
As shown in Figure~\ref{fig:method_blocks} (d), the divide and multiply feed-forward network (DMFN) omits the activation function in previous Transformer FFNs, inspired by the simple gate in CNN-based NAFNet~\citep{chen2022simple}. We assert that DeblurDiNAT remains a robust model because the Leaky ReLU in the encoder path ensures non-linearity in latent features, and the feature fusion modules with GELU propagate non-linear features cross different network levels. From a layer normalized input feature $\mathbf{X}\in\mathbb{R}^{\hat{H}\times\hat{W}\times\hat{C}}$, DMFN first applies point-wise convolutions to expand the feature channels by a factor $\gamma=2$. After that, $3\times3$ depth-wise convolutions are utilized to aggregate spatial context, generating $\mathbf{X}_0\in\mathbb{R}^{\hat{H}\times\hat{W}\times2\hat{C}}$. Next, the intermediate feature map $\mathbf{X}_0$ is equally divided along the channel dimension, yielding
$\mathbf{X}_1$ and $\mathbf{X}_2$, so that their element-wise multiplication result is of shape $\mathbb{R}^{\hat{H}\times\hat{W}\times\hat{C}}$.
The overall process of DMFN is formulated as:
\begin{equation}
  \begin{aligned} \label{eq1}
  \hat{\mathbf{X}} &= W_d^1W_p^1\mathbf{X}\odot W_d^2W_p^2\mathbf{X},
  \end{aligned}
\end{equation}
where $W_p^{(\cdot)}$ and $W_d^{(\cdot)}$ denote the convolution with filter size of $1\times1$ and $3\times3$. 

\subsection{Lightweight Dual-Stage Feature Fusion}

\begin{figure*}[tb]
  \centering
  \includegraphics[width=0.98\linewidth, trim = 0.5cm 14.8cm 10.4cm 0.50cm]{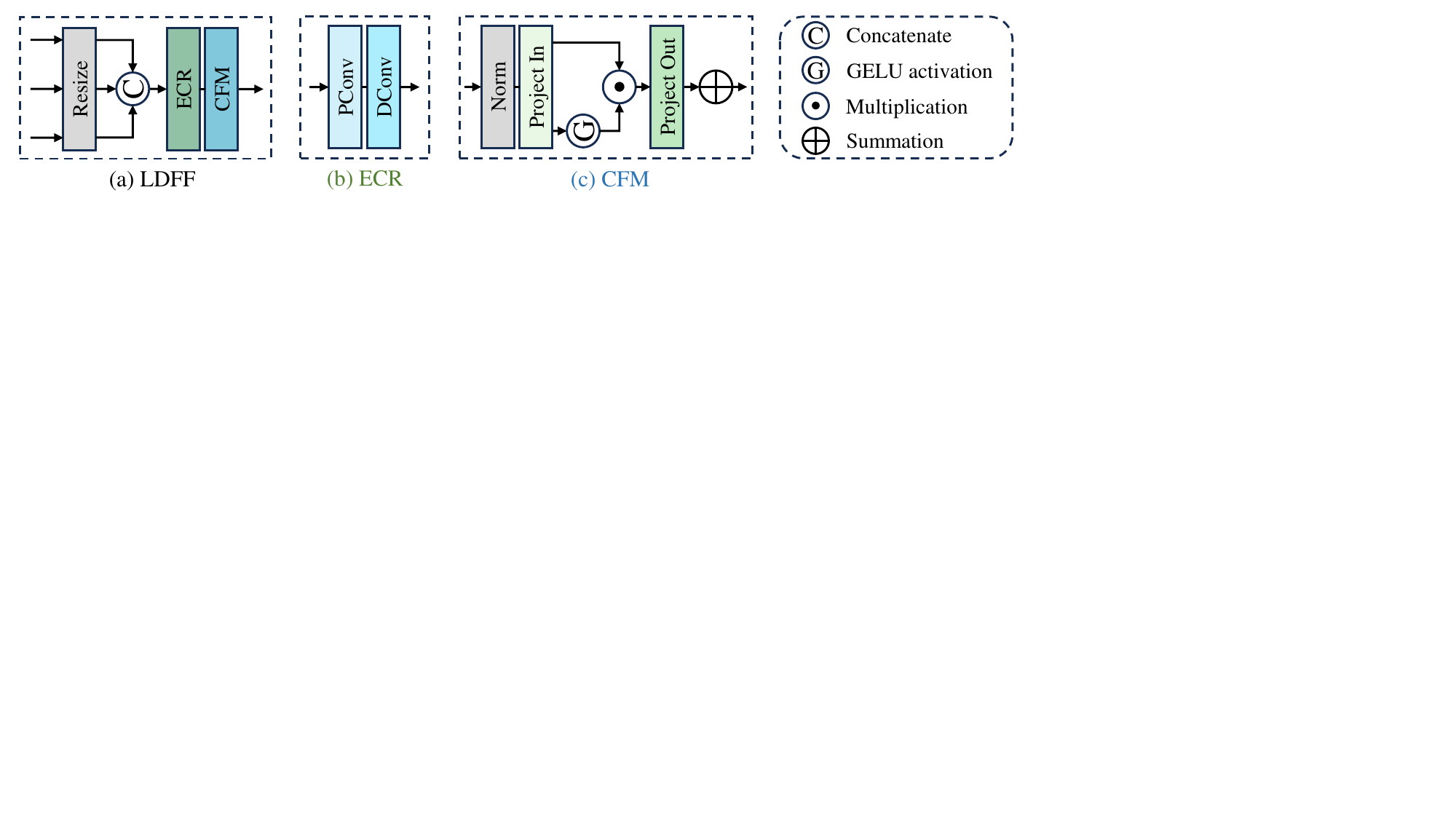}

  \caption{Structures of (b) lightweight dual-stage feature fusion (LDFF), composed of (b.1) efficient channel reduction (\textcolor{darkgreen}{ECR}) and (b.2) complementary feature mixer (\textcolor{darkblue}{CFM}).
  }\label{fig:method_ldff}
\end{figure*}

As shown in Figure~\ref{fig:method_ldff} (a)-(c), LDFF first resizes input features from different encoder levels, and concatenates them along the channel dimension. LDFF is composed of two stages, efficient channel reduction (ECR) and complementary feature mixing (CFM). In ECR, a $1\times1$ convolution is followed by a $3\times3$ depth-wise convolution. Then, CFM splits a normalized feature map into two complementary features by convolutional layers, passing through two parallel branches, one of which is gated by GELU activation. Finally, the two complementary features are merged by element-wise multiplications and additional convolutions. The two multi-scale LDFF blocks in DeblurDiNAT are formulated as:
\begin{equation}
  \begin{aligned} \label{eq2}
  &\mathrm{LDFF}_1^{out} = \mathrm{LDFF}_1 (\mathrm{EB}_1^{out}, (\mathrm{EB}_2^{out})^{\times2}, (\mathrm{EB}_3^{out})^{\times4}), \\
  &\mathrm{LDFF}_2^{out} = \mathrm{LDFF}_2 ((\mathrm{EB}_1^{out})^{\times \frac{1}{2}}, \mathrm{EB}_2^{out}, (\mathrm{EB}_3^{out})^{\times2}),\\
\end{aligned}
\end{equation}
where $\mathrm{LDFF}_i^{out}$ and $\mathrm{EB}_i^{out}$ denote the output features of the $\ith$th LDFF and
the $\ith$th encoder block;
$\times2$ and $\times4$ represent upscaling by $2$ and $4$;
$\times \frac{1}{2}$ means downscaling by $2$.
LDFF for same-scale feature fusion is of a similar architecture without the resizing procedure.

\section{Experiments and Analysis}\label{sec:experiment}
\noindent \textbf{Datasets.}
The GoPro dataset~\citep{nah2017deep} contains a variety of outdoor scenes, while the HIDE dataset~\citep{shen2019human} is distinct for its focus on human images. Both datasets share similar data collection methods. RealBlur-R~\citep{rim2020real} and RealBlur-J~\citep{rim2020real} are closer to real-world blurred images, with most images captured in low-light conditions. RWBI~\citep{zhang2020deblurring} is another real-world dataset but does not include ground-truth images. Since the focus of this paper is generalization capabilities and perceptual performance, DeblurDiNAT was trained only on the GoPro training set and was directly applied to HIDE, RealBlur-R, RealBlur-J, and RWBI test sets.

\noindent \textbf{Implementation Details.}
We propose DeblurDiNAT-S (Small) and -L (Large). 
The number of Transformer blocks in DeblurDiNAT-S are $[4, 6, 8]$, 
attention heads are $[2, 4, 8]$, and the number of channels are $[64, 128, 256]$ at each level.
In the large network, one more residual block is appended to each encoder;
the number of Transformer blocks increment to $[6, 12, 18]$.
We trained both models using Adam optimizer with the initial learning rate of $2\times10^{-4}$
which was gradually decayed to $10^{-7}$ by the cosine annealing strategy. 
The batch size was set to $8$.
DeblurDiNAT-L converged after training for 4000 epochs with a patch size of $256\times256$ 
and another 2000 epochs with a patch size of $496\times496$, on eight NVIDIA A100 GPUs.

\noindent \textbf{Evaluation Metrics.}
Following previous work~\citep{whang2022deblurring}, FID$\downarrow$ (Fréchet Inception Distance)~\citep{heusel2017gans}, KID$\downarrow$ (Kernel Inception Distance)~\citep{binkowski2018demystifying}, LPIPS$\downarrow$~\citep{zhang2019deep} and NIQE$\downarrow$~\citep{mittal2012making} are employed as perceptual metrics. We adopt the Shift-Tolerant Perceptual Similarity Metric (ST-LPIPS$\downarrow$, {\bftab abbreviated as S-LP in Tables})~\citep{ghildyal2022stlpips} as an alternative to LPIPS~\citep{zhang2018unreasonable}. FID and KID are normalized to a range of $[0, 1]$.We also evaluate image quality by distortion metrics such as PSNR$\uparrow$ and SSIM$\uparrow$ for a fair comparison. 

\subsection{Comparisons with State-of-the-art Methods}
In this paper, the deblurring networks with parameter count larger than 40M are considered as {\bftab non-compact networks}, labeled with $^{\dag}$ in Tables \ref{tab:perception} and \ref{tab:distortion}. Since the focus of this paper is generalization and visual fidelity, we evaluate the performance of GoPro-trained models on unknown domains, and provide the perceptual metric results in Table~\ref{tab:perception} and deblurred images in Figure. We also report distortion metrics in Table~\ref{tab:distortion}.

\subsubsection{Quantitative Comparisons}

\begin{table*}[tb]
  \setlength{\tabcolsep}{3.pt}
  \centering
  \caption{
  Perceptual metric scores of GoPro-trained~\citep{nah2017deep} models on HIDE~\citep{shen2019human}, RealBlur-J and RealBlur-R~\citep{rim2020real} test sets. Best results are {\bftab highlighted}, the second best results are \ul{underlined}, and the third highest results are \ull{double-underlined}. In this paper, the deblurring networks with parameter count more than 40M are considered as non-compact models marked with $^{\dag}$.}\label{tab:perception}
  \begin{tabular}{l c cccc cccc cccc}
    \toprule
    \multirow{2}{*}{Models} & \#P. & \multicolumn{4}{c}{HIDE} & \multicolumn{4}{c}{RealBlur-J} & \multicolumn{4}{c}{RealBlur-R} \\
    \cmidrule(lr){3-6} \cmidrule(lr){7-10} \cmidrule(lr){11-14}
    \rowcolor{Gray}
    &(M) & FID & KID & S-LP & NIQE & FID & KID & S-LP & NIQE & FID & KID & S-LP & NIQE\\
    \midrule
    MIMO-Unet{\Rplus}~\textcolor{red}{\citeyear{cho2021rethinking}} 
    &\ul{16.1} &0.178 &0.062 &0.00 &\ull{3.24} &0.965 &0.770 &0.142 &3.99 &0.637 &0.273 &0.034 &5.34 \\  
    \rowcolor{Gray}
    MPRNet~\textcolor{red}{\citeyear{zamir2021multi}} 
    &20.1 &0.168 &0.054 &0.070 &3.46 &0.606 &0.333 &0.103 &3.97 &0.550 &0.243 &0.031 &5.44 \\
    NAFNet$^{\dag}$~\textcolor{red}{\citeyear{chen2022simple}} 
    &67.9 &\ull{0.137} &\ull{0.043} &0.058 &\ul{3.22} &0.699 &0.435 &0.114 &\bftab3.77 &0.467 &0.133 &0.030 &5.36 \\
    \rowcolor{Gray}
    UFPNet$^{\dag}$~\textcolor{red}{\citeyear{fang2023self}} 
    &80.3 &0.150 &0.065 &\bftab0.051 &\bftab3.13 &\bftab0.460 &\bftab0.255 &\bftab0.081 &4.07 &\bftab0.402 &0.135 &\ul{0.025} &\bftab5.17 \\
    \midrule
    Restormer~\textcolor{red}{\citeyear{zamir2022restormer}} 
    &26.1 &0.150 &0.052 &0.063 &3.42 &0.576 &0.318 &0.099 &3.92 &0.436 &0.166 &\ull{0.026} &\ull{5.23} \\
    \rowcolor{Gray}
    Uformer-B$^{\dag}$~\textcolor{red}{\citeyear{wang2022uformer}} 
    &50.9 &0.141 &0.052 &0.065 &3.40 &\ull{0.515} &\ul{0.261} &\ull{0.088} &3.86 &\ull{0.421} &\ull{0.129} &\bftab0.024 &5.28 \\
    FFTformer~\textcolor{red}{\citeyear{kong2023efficient}} 
    &\ull{16.6} &0.139 &0.051 &0.054 &3.29 &0.907 &0.721 &0.128 &3.90 &0.500 &0.157 &0.027 &5.24 \\
    \rowcolor{Gray}
    LoFormer-L$^{\dag}$~\textcolor{red}{\citeyear{mao2024loformer}} 
    &49.0 &\ul{0.125} &\ul{0.040} &\ul{0.052} &3.30 &0.672 &0.416 &0.110 &\ul{3.79} &0.454 &0.157 &0.027 &\bftab5.17 \\
    \midrule
    \midrule
    {\bftab DeblurDiNAT-S} 
    &\bftab9.1 &0.143 &0.045 &0.063 &3.43 &0.578 &0.315 &0.095 &\ull{3.83} &0.458 &\ul{0.122} &0.027 &\ul{5.21} \\
    \rowcolor{Gray}
    {\bftab DeblurDiNAT-L} 
    &\ul{16.1} &\bftab0.121 &\bftab0.034 &\ull{0.053} &3.49 &\ul{0.509} &\ull{0.265} &\ul{0.084} &\ul{3.79} &\ul{0.417} &\bftab0.104 &\ul{0.025} &\bftab5.17 \\
    \bottomrule
  \end{tabular}
\end{table*}

\begin{table*}[tb]
  \centering
  \caption{
  Distortion metric comparisons on HIDE~\citep{shen2019human} and RealBlur~\citep{rim2020real} datasets. $^{\star}$ means that we directly use the image quality values reported in the original paper~\citep{fang2023self}.}\label{tab:distortion}
  \begin{tabular}{l c c cc cc cc}
    \toprule
    \multirow{2}{*}{Models} & Param. & Average & \multicolumn{2}{c}{HIDE} & \multicolumn{2}{c}{RealBlur-J} & \multicolumn{2}{c}{RealBlur-R}\\
    \cmidrule(lr){4-5} \cmidrule(lr){6-7} \cmidrule(lr){8-9}
    \rowcolor{Gray}
    & (M) & PSNR & PSNR & SSIM & PSNR & SSIM & PSNR & SSIM\\
    \midrule
    MIMO-Unet{\Rplus} 
    &\ul{16.1} &31.05 & 29.99 &0.930 & 27.63 &0.837 & 35.54 &0.947\\
    \rowcolor{Gray}
    MPRNet 
    &20.1 &31.88 & 30.96 &0.939 & 28.70 &0.873 & 35.99 &0.952\\
    NAFNet$^{\dag}$ 
    &67.9  &31.87 &31.32 &0.943 & 28.32 &0.857& 35.97 &0.951\\
    \rowcolor{Gray}
    UFPNet$^{\dag \star}$ 
    &80.3 &\bf32.62 &\bftab31.74 &\ull{0.947}  &\bftab29.87 &\ull{0.884} &\bftab36.25 &0.947\\
    \midrule
    Restormer 
    &26.1 &32.12 & 31.22 &0.942 &28.96 &0.879 & \ul{36.19} &\bftab0.957\\
    \rowcolor{Gray}
    Uformer-B$^{\dag}$ 
    &50.9 &32.06 & 30.90 &\bftab0.953 & \ul{29.09} &\bftab0.886 & \ul{36.19} &\ul{0.956}\\
    FFTformer &16.6 &31.75 & \ull{31.62} &0.946 & 27.75 &0.853 & 35.87 &0.953\\
    \rowcolor{Gray}
    LoFormer-L$^{\dag}$ &49.0 &\ull{32.16} &\ul{31.86} &\ul{0.949} &28.53 &0.863 &36.09 &\ull{0.955}\\
    \midrule
    \midrule
    {\bftab DeblurDiNAT-S} &{\bftab 9.1} &31.86 & 30.92 &0.939 & 28.73 &0.874 & 35.93 &0.953\\
    \rowcolor{Gray}
    {\bftab DeblurDiNAT-L} &\ul{16.1} &\ul{32.18} & 31.47 &0.944 & \ull{28.98} &\ul{0.885} & \ull{36.09} &\ull{0.955}\\
    \bottomrule
  \end{tabular}
\end{table*}

Tables \ref{tab:perception} demonstrates that our proposed DeblurDiNAT-L achieves comparable or superior perceptual performance to state-of-the-art methods across multiple metrics while utilizing favorably fewer parameters. Specifically, compared to the other compact networks MIMO-Unet{\Rplus}~\citep{cho2021rethinking}, MPRNet~\citep{zamir2021multi}, Restormer~\citep{zamir2022restormer} and FFTformer~\citep{kong2023efficient}, our DeblurDiNAT-L achieves {\bftab significantly improved deblurring performance}. Compared to the closest competitor, UFPNet~\citep{fang2023self}, DeblurDiNAT-L delivers on-par performance with {\bftab only $\mathbf{20\%}$ of the parameters}. While Figure~\ref{fig:intro_compare_metrics} has implied that PSNR is not a good option to evaluate perceptual performance, we report distortion metrics in Table~\ref{tab:distortion}. Compared to recent deblurring Transformers such as FFTformer~\citep{kong2023efficient} and LoFormer~\citep{mao2024loformer}, Table \ref{tab:distortion} reports that DeblurDiNAT-L exhibits competitive distortion metric scores on real-world datasets. A detailed analysis of quantitative results on each dataset is provided as follows.

\noindent \textbf{HIDE.} 
The fine facial textures in the human centric images elevate NIQE values~\citep{zvezdakova2019barriers}. We still report NIQE on the human centric dataset HIDE~\citep{shen2019human} for thorough comparisons. As shown in Table~\ref{tab:perception}, the proposed method achieves the state-of-the-art FID and KID, while delivering comparable ST-LPIPS to UFPNet~\citep{fang2023self}. It proves that our model demonstrates great visual performance and generalizes very well to unknown images with human beings.

\noindent \textbf{RealBlur-J.}
Table~\ref{tab:perception} reports that our DeblurDiNAT-L ranks right after UFPNet~\citep{fang2023self} in terms of FID and ST-LPIPS while with $79.95\%$ fewer parameters. Notably, our method overperforms UFPNet~\citep{fang2023self} by 0.28 as rated by NIQE. It verifies that the proposed approach restores real-world blurred images favorably close to human perception.

\noindent \textbf{RealBlur-R.}
Table~\ref{tab:perception} shows that DeblurDiNAT-L delivers the state-of-the-art KID and NIQE, followed by our small version DeblurDiNAT-S. Compared to UFPNet~\citep{fang2023self}, DeblurDiNAT-S achieves competitive KID, ST-LPIPS and NIQE while with only $11.33\%$ parameters on RealBlur-R~\citep{rim2020real}. It suggests the superiority of our method in handling real-world blurs with low-light illuminations.

\subsubsection{Qualitative Comparisons}
\begin{figure*}[tb]
  \centering
    \begin{subfigure}{\linewidth}
        \centering
          \includegraphics[width=0.98\linewidth, trim = 0.05cm 13.6cm 18.7cm 0.05cm]{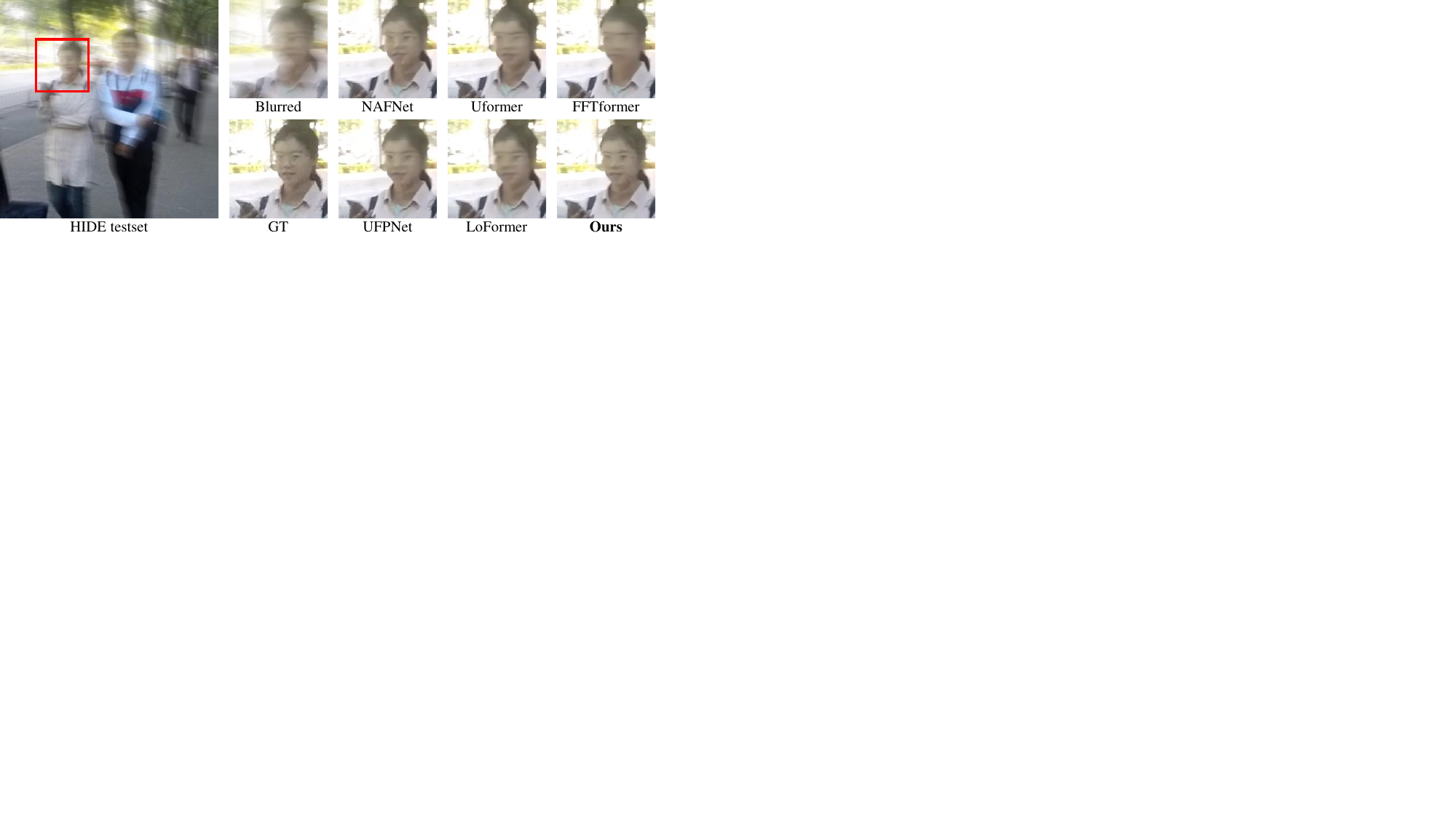}
    \end{subfigure}

    \vspace{0.3em} 

    \begin{subfigure}{\linewidth}
        \centering
          \includegraphics[width=0.98\linewidth, trim = 0.05cm 13.6cm 18.7cm 0.05cm]{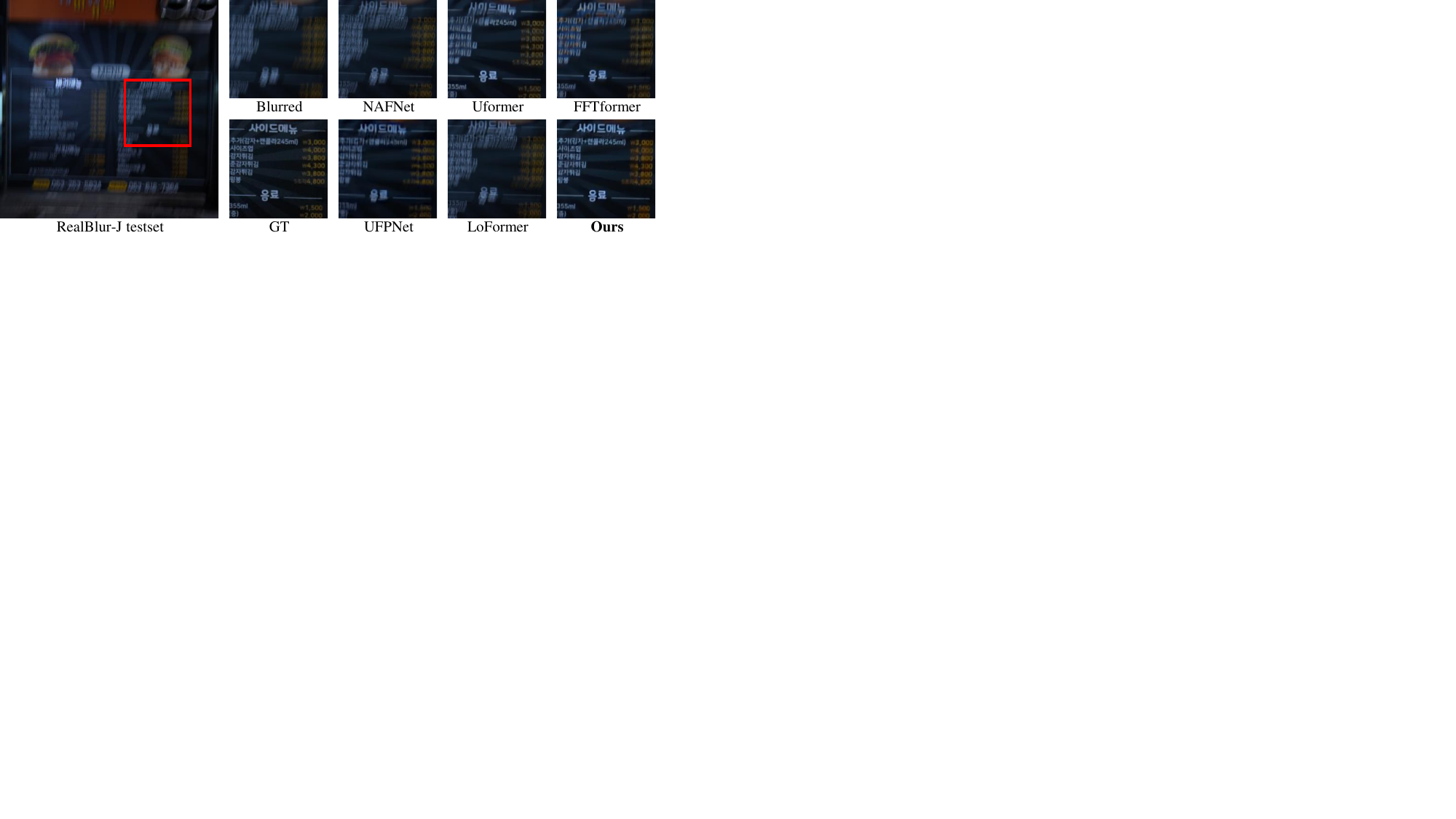}
    \end{subfigure}
    
    \vspace{0.3em} 
    
    \begin{subfigure}{\linewidth}
        \centering
          \includegraphics[width=0.98\linewidth, trim = 0.05cm 13.6cm 18.7cm 0.05cm]{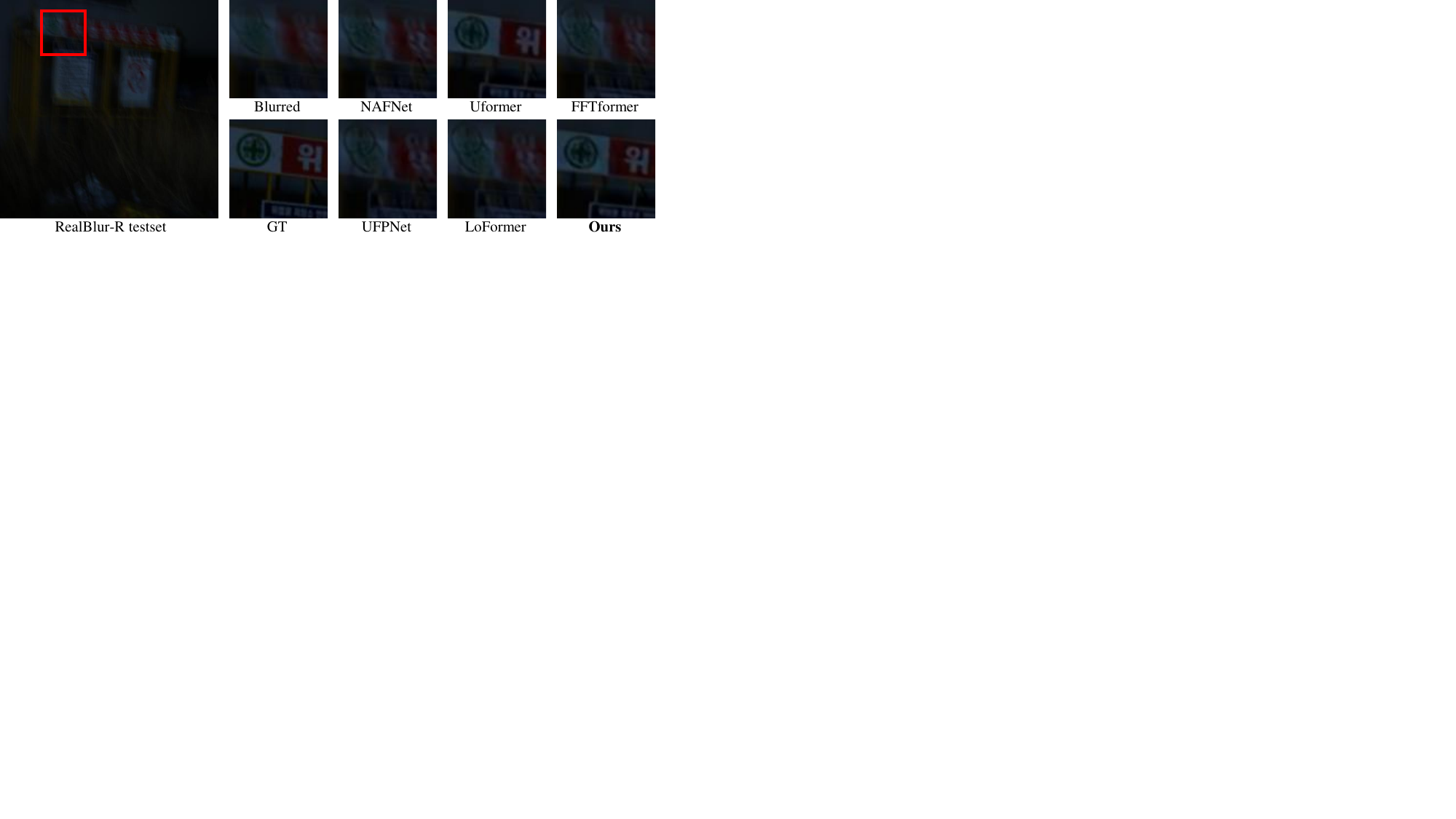}
    \end{subfigure}
  \caption{Deblurred images of GoPro-trained models on the unknown HIDE~\citep{shen2019human}, RealBlur-J and -R~\citep{rim2020real} datasets. Best viewed with zoom and high brightness for details.}\label{fig:experiment_image}
\end{figure*}

\begin{figure*}[tb]
  \centering
    \begin{subfigure}{\linewidth}
        \centering
          \includegraphics[width=0.98\linewidth, trim = 0.05cm 13.6cm 18.7cm 0.05cm]{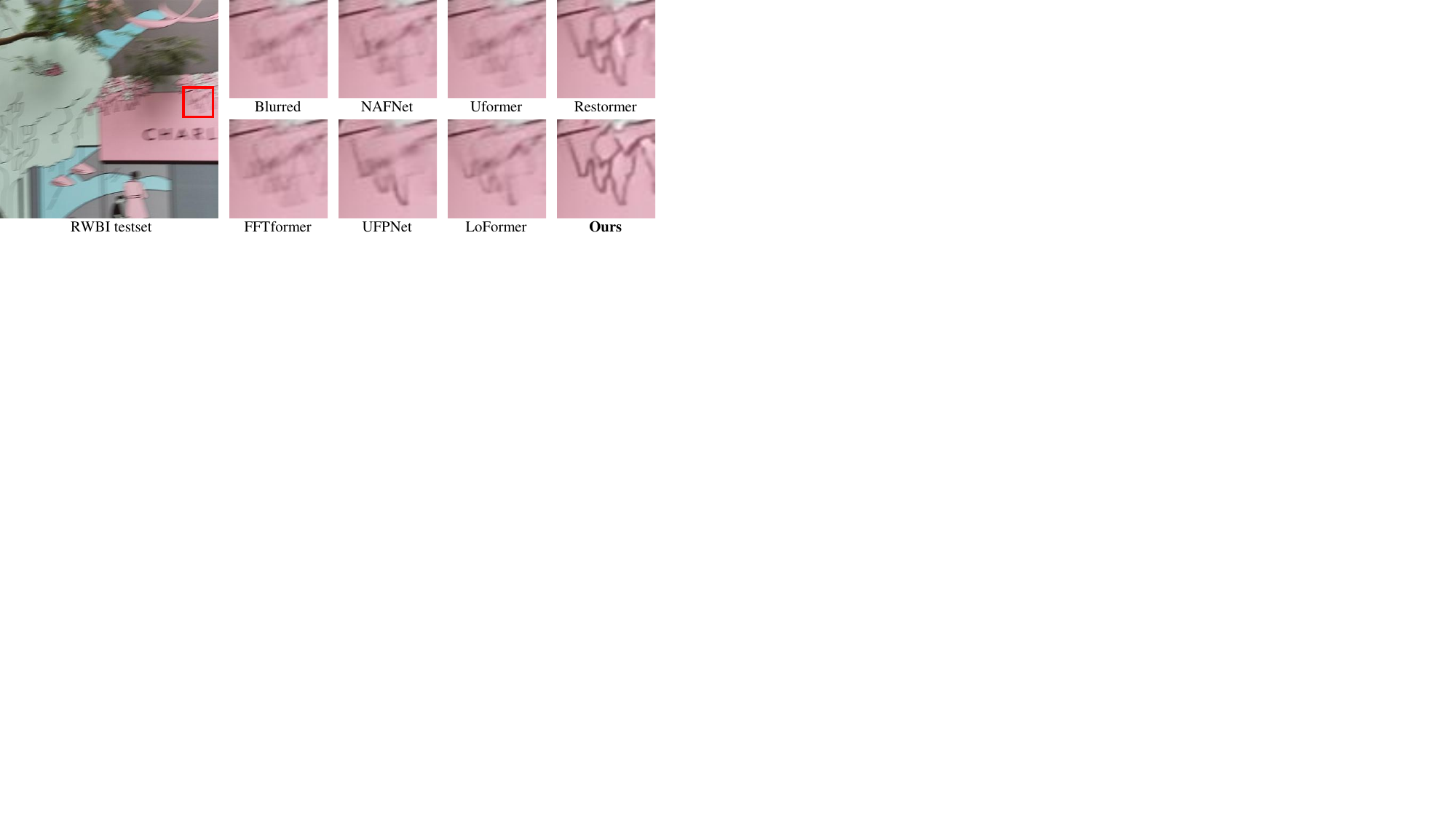}
    \end{subfigure}

    \vspace{0.3em} 

    \begin{subfigure}{\linewidth}
        \centering
          \includegraphics[width=0.98\linewidth, trim = 0.05cm 13.6cm 18.7cm 0.05cm]{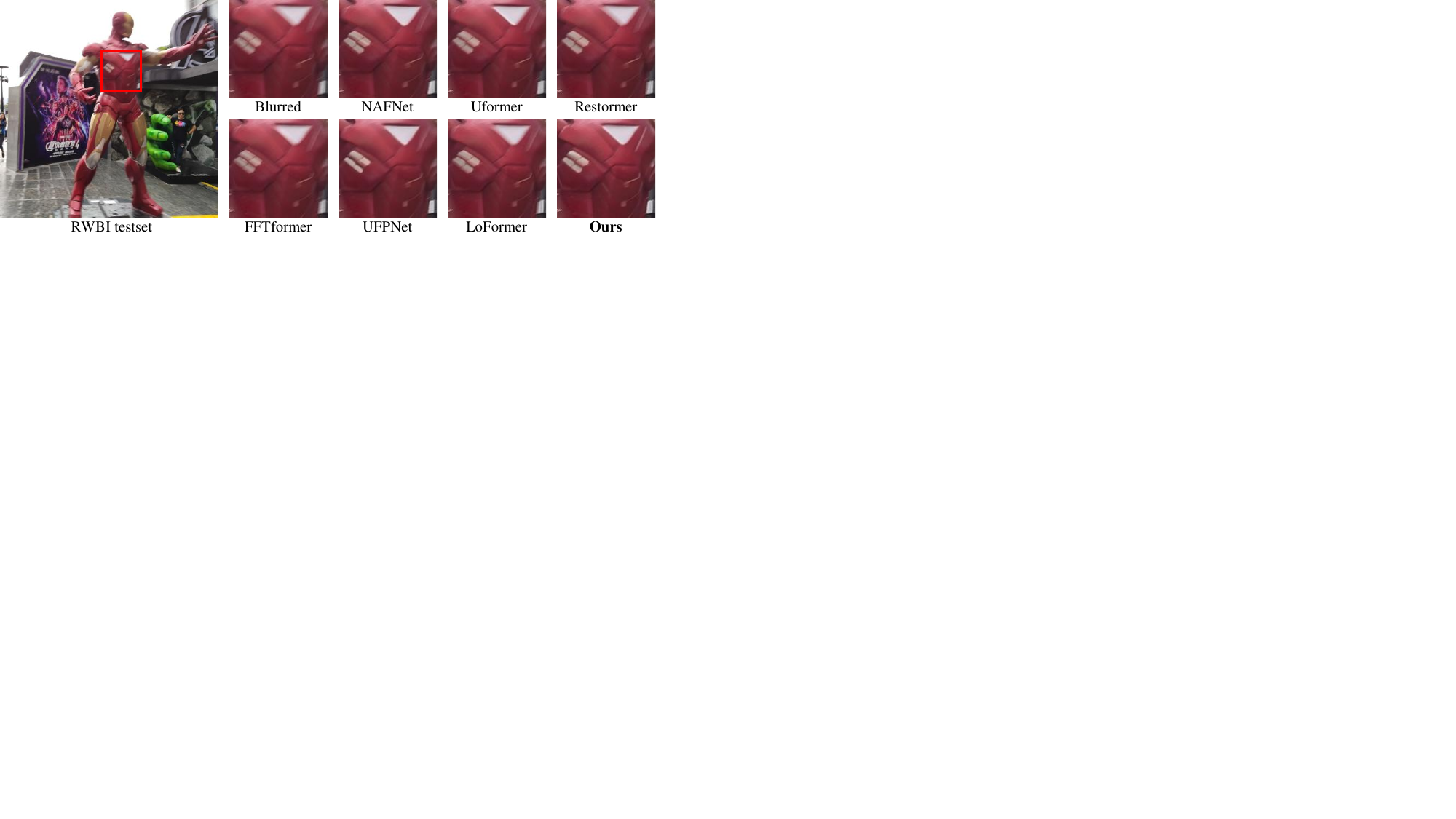}
    \end{subfigure}
    
    \vspace{0.3em} 

    \begin{subfigure}{\linewidth}
        \centering
          \includegraphics[width=0.98\linewidth, trim = 0.05cm 13.6cm 18.7cm 0.05cm]{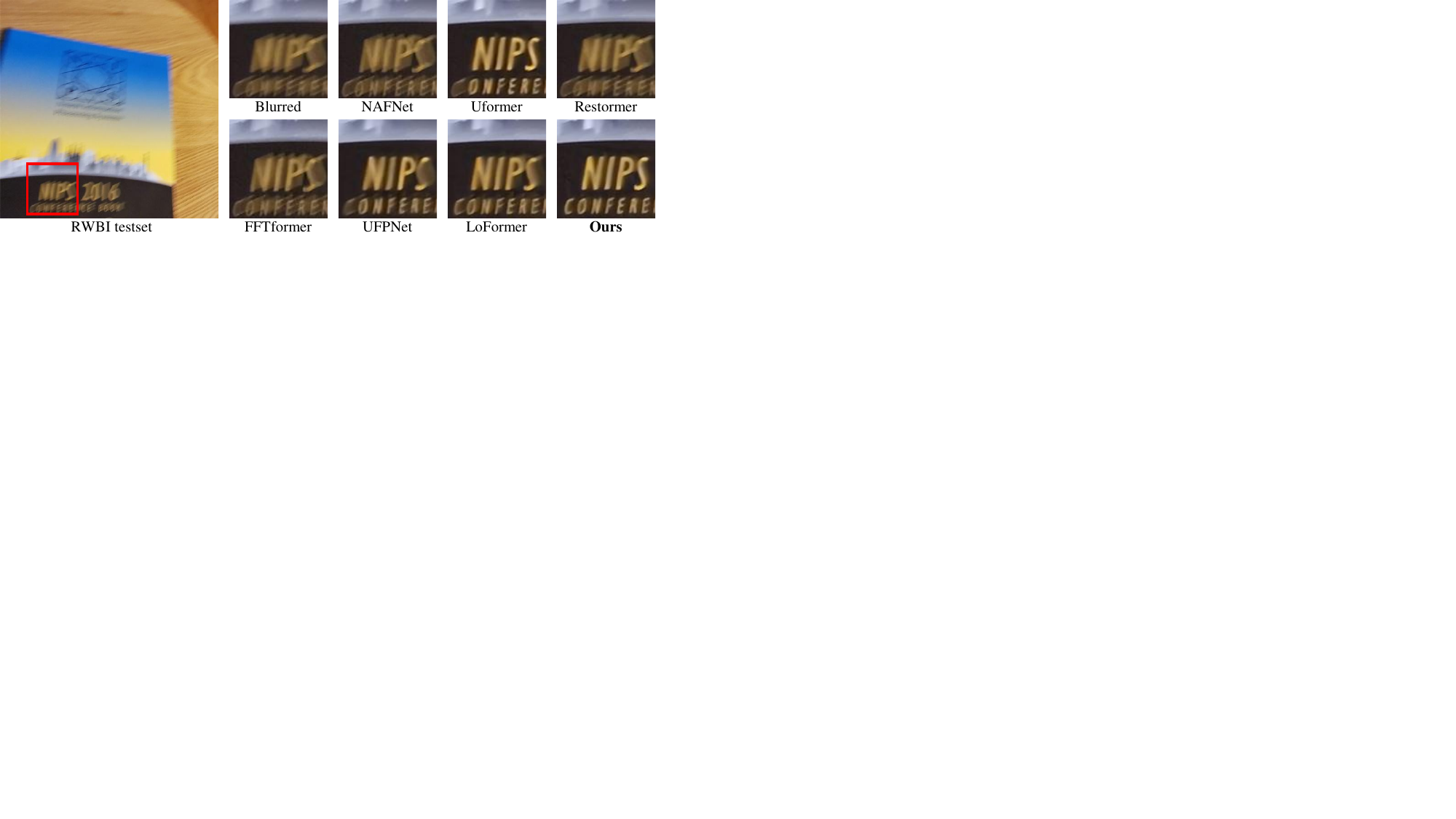}
    \end{subfigure}
  \caption{Deblurred images of GoPro-trained models on the unknown RWBI dataset~\citep{zhang2020deblurring}.}\label{fig:experiment_rwbi_img}
\end{figure*}

As illustrated in Figure~\ref{fig:experiment_image}, our proposed model, DeblurDiNAT-L, outperforms recent deblurring networks by generating noticeably clearer human faces on the HIDE dataset~\citep{shen2019human}. On real-world datasets such as RealBlur-J and RealBlur-R~\citep{rim2020real}, DeblurDiNAT-L demonstrates superior performance by producing cleaner and sharper text characters, particularly under challenging low-light conditions. These results highlight the generalization capabilities and visual fidelity of our model compared to existing approaches. Notably, this performance is achieved with a minimal number of parameters.

\noindent \textbf{RWBI.}
Figure~\ref{fig:experiment_rwbi_img} demonstrates that the proposed DeblurDiNAT-L effectively restores significantly sharper streaks compared to existing methods on RWBI~\citep{zhang2020deblurring}. For instance, in the middle row (iron man) of Figure~\ref{fig:experiment_rwbi_img}, the forward-slash-shaped line at the top right is restored with remarkable clarity.

\subsection{Ablation Study}
We conduct ablation experiments to evaluate each component of DeblurDiNAT. Models were trained on the GoPro dataset~\citep{nah2017deep} with a patch size of $128\times128$ for 3000 epochs only. Given that tradeoff between reconstruction accuracy and perceptual quality exists~\citep{blau20182018}, both quantitative comparisons and qualitative results are provided to thoroughly demonstrate the effectiveness of our methods.

\begin{table}[tb]
  \centering
  \caption{Effects of different feature fusion strategies, AFF~\citep{cho2021rethinking} and our LDFF. Exploration on combinations of different dilation factors, the lower bound only, the upper bound only and the hybrid.}
  \begin{tabular}{c  c  c  c  c  c  c  c}
    \toprule
    \multicolumn{2}{c}{Feature Fusion} &\multicolumn{2}{c}{Dilation Factor} &\multicolumn{2}{c}{HIDE} & \multicolumn{2}{c}{RealBlur-J} \\
    \cmidrule(lr){1-2} \cmidrule(lr){3-4} \cmidrule(lr){5-6} \cmidrule(lr){7-8}
    \rowcolor{Gray}
    AFF &\bftab LDFF & $\delta=1$ &$\delta=\lfloor\frac{n}{k}\rfloor$ &S-LP$\downarrow$ &PSNR$\uparrow$ &S-LP$\downarrow$ &PSNR$\uparrow$\\
    \midrule
    \checkmark&  & &\checkmark           &0.084 &29.76 &\bftab0.097 &28.51\\
    &\checkmark&           &\checkmark &0.082 &29.86 &0.098 &\bftab28.55\\
    \rowcolor{Gray}
    &\checkmark&\checkmark &           &0.085 &29.47 &0.101 &28.34\\
    &\checkmark&\checkmark &\checkmark &\bftab0.078 &\bftab29.94 &0.098 &28.54\\
  \bottomrule
  \end{tabular}
  \label{tab:ablation_ff_df}
\end{table}

\begin{figure}[tb]
  \centering
  \includegraphics[width=0.98\linewidth, trim = 0.13cm 15.13cm 11.88cm 0.13cm]{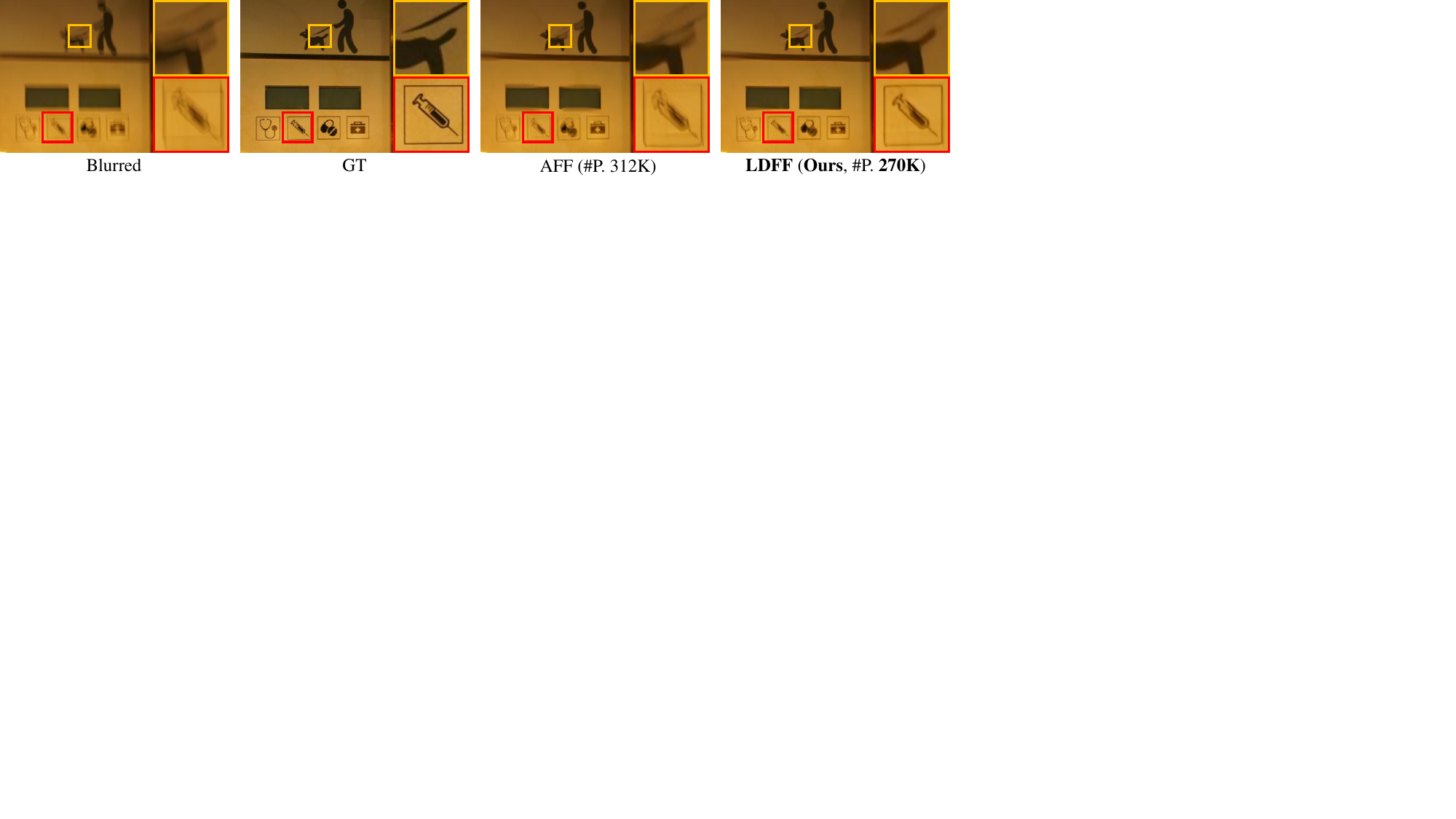}
     \caption{The proposed feature fusion method improves visual quality with $13.46\%$ smaller module size.}\label{fig:ablation_fusion}
\end{figure}

\begin{figure}[tb]
  \centering
  \includegraphics[width=0.98\linewidth, trim = 0.13cm 13.69cm 11.88cm 0.13cm]{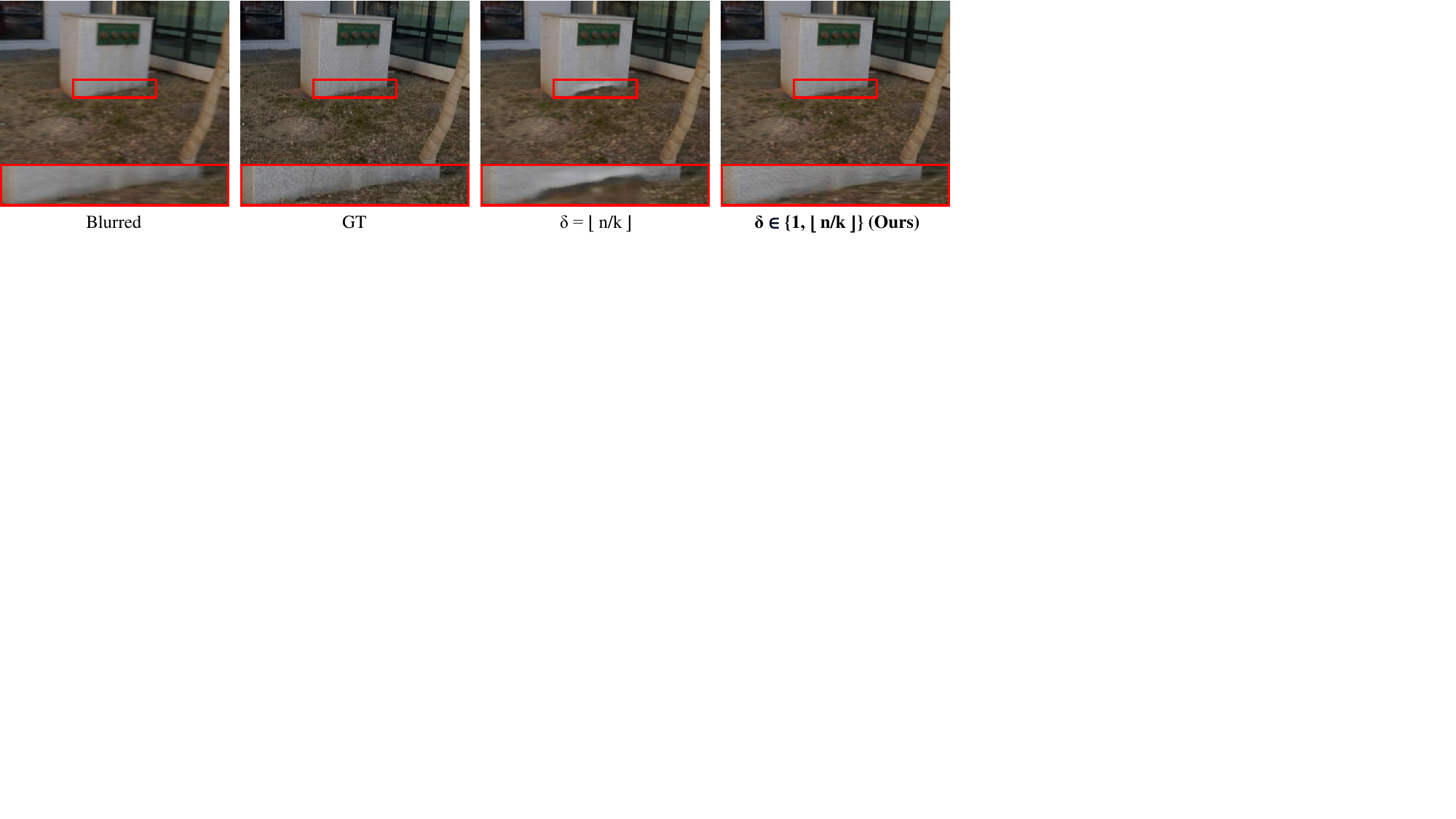}
     \caption{The presented alternating structure with hybrid dilation factors reduces artifacts in small regions.}\label{fig:ablation_delta}
\end{figure}

\begin{table}[tb]
  \centering
  \caption{Investigation on the proposed channel-aware self-attention (CA-SA) with a local cross channel learner (LCCL), and comparison between GDFN~\citep{zamir2022restormer} and the presented DMFN. Hue Distance denotes the average difference between the hue of deblurred images and ground truth in the HSV color space.
  }
  \begin{tabular}{c  c  c  c  c  c  c  c  c}
    \toprule
    \multicolumn{2}{c}{Self-Attention} &\multicolumn{2}{c}{FFN} &\multicolumn{2}{c}{HIDE} & \multicolumn{2}{c}{RealBlur-J} &Hue Distance\\
    \cmidrule(lr){1-2} \cmidrule(lr){3-4} \cmidrule(lr){5-6} \cmidrule(lr){7-8}
    \rowcolor{Gray}
    SA &\bftab CA-SA &GDFN & \bftab DMFN &S-LP$\downarrow$ &PSNR$\uparrow$ &S-LP$\downarrow$ &PSNR$\uparrow$ &\%\\
    \midrule
    \checkmark &           &\checkmark &           &0.078 &29.94 &0.098 &28.54 &0.65\\
    \rowcolor{Gray}
               &\checkmark &\checkmark &           &0.080 &29.97 &0.096 &28.61 &\bftab0.58\\
    \checkmark &           &           &\checkmark &0.079 &29.93 &0.096 &28.53 &0.70\\
    \rowcolor{Gray}
               &\checkmark &           &\checkmark &\bftab0.077 &\bftab29.99 &\bftab0.096 &\bftab28.59 &\bftab0.58\\
  \bottomrule
  \end{tabular}\label{tab:ablation_sa_ffn}
\end{table}

\begin{figure}[tb]
  \centering
  \includegraphics[width=0.98\linewidth, trim = 0.13cm 12.9cm 11.83cm 0.13cm]{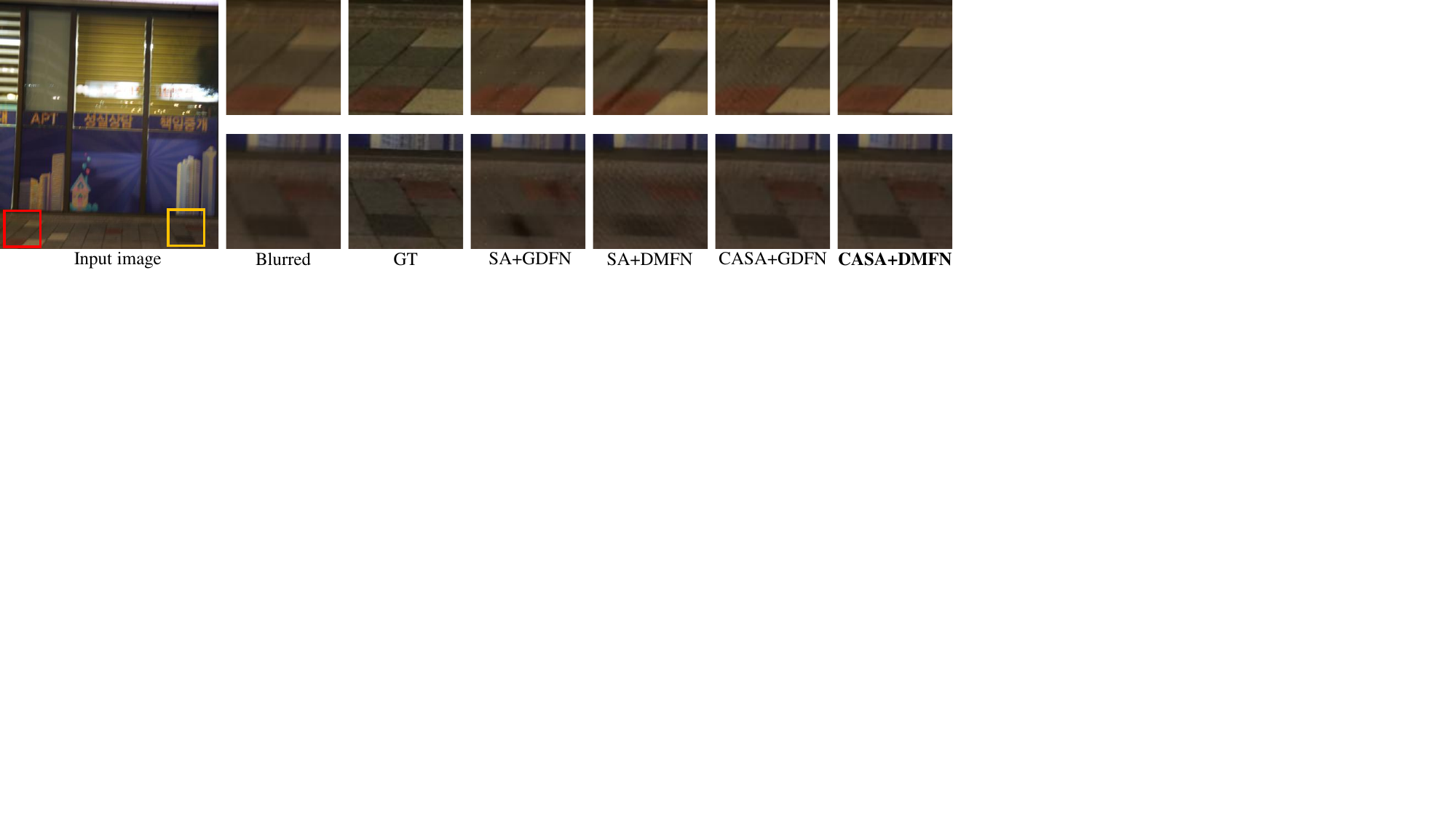}
     \caption{The proposed methods CA-SA and DMFN generate road textures closer to the ground truth.}\label{fig:ablation_casa_dmfn}
\end{figure}

\noindent \textbf{Feature Fusion.} We first compare the proposed LDFF with the previous AFF in MIMO-Unet~\citep{cho2021rethinking}. As shown in Table~\ref{tab:ablation_ff_df}, LDFF improves both perceptual and distortion metrics on HIDE, and achieves similar ST-LPIPS and noticeably higher PSNR on RealBlur-J. Figure~\ref{fig:ablation_fusion} shows that our method produces sharper edges, verifying its superior utilization of multi-scale features. Notably, the lightweight dual-stage feature fusion (LDFF) in DeblurDiNAT contains only {\bftab 270K} parameters, significantly fewer than the 312K parameters required by the feature fusion methods in MIMO-Unet~\citep{cho2021rethinking}.

\noindent \textbf{Dilation Factor.} We experiment with different combinations of dilation factors, $\delta=1$, $\delta=\lfloor\frac{n}{k}\rfloor$, and the hybrid $\delta \in \left\{1, \lfloor\frac{n}{k}\rfloor\right\}$. As shown in Table~\ref{tab:ablation_ff_df}, the presented structure incorporating both small and big dilation factor obtains the best overall performance. Furthermore, Figure~\ref{fig:ablation_delta} suggests that the introduction of small dilation factors effectively reduces artifacts in fine areas, resulting in smoother restorations.

\noindent \textbf{Channel Aware Self Attention.} We evaluate the effectiveness of CA-SA by comparing it with self-attention that excludes the Local Cross-Channel Learner (LCCL). As presented in Table~\ref{tab:ablation_sa_ffn}, CA-SA achieves consistent quantitative improvements across various metrics on multiple datasets. Additionally, the hue analysis in Table~\ref{tab:ablation_sa_ffn} reveals that CA-SA produces colors much closer to the reference, highlighting the LCCL module's ability to enhance the network’s cross-channel learning, particularly for color information. Figure~\ref{fig:ablation_casa_dmfn} illustrates that CA-SA noticeably improves the visual quality of deblurred images.

\noindent \textbf{Feed-Forward Network.} We compare the proposed Divide-and-Multiply Feed-Forward Network (DMFN) with the Gated FFN (GDFN)~\citep{zamir2022restormer} in DeblurDiNAT. Since non-linearity has been ensured in the encoder path and feature fusion modules, DeblurDiNAT eliminates the need for GELU activation in FFN. As shown in Table~\ref{tab:ablation_sa_ffn}, DMFN delivers competitive performance compared to GDFN, showcasing its efficiency. Figure~\ref{fig:ablation_casa_dmfn} reveals that our method effectively captures boundaries between distinct regions.

\section{Conclusion}
In this work, we propose DeblurDiNAT, a novel deblurring Transformer that combines strong generalization capabilities and impressive visual fidelity with a minimal model size. First, an alternating structure with different dilation factors is introduced to effectively estimate blurred patterns of varying magnitudes. Second, a local cross-channel learner adaptively modulates self-attention outputs, enhancing channel-wise learning for attributes such as color information. To enable efficient feature propagation, a simple yet effective feed-forward network is employed, replacing traditional Transformer designs. Additionally, a lightweight dual-stage feature fusion method ensures non-linearity and aggregates multi-scale features for improved performance. Comprehensive experiments demonstrate that DeblurDiNAT achieves comparable quantitative results and significantly superior qualitative performance compared to the state-of-the-art UFPNet~\citep{fang2023self}, while utilizing only $20\%$ of the parameters.




\bibliography{tmlr}

\begin{thebibliography}{49}
\providecommand{\natexlab}[1]{#1}
\providecommand{\url}[1]{\texttt{#1}}
\expandafter\ifx\csname urlstyle\endcsname\relax
  \providecommand{\doi}[1]{doi: #1}\else
  \providecommand{\doi}{doi: \begingroup \urlstyle{rm}\Url}\fi

\bibitem[Bahat et~al.(2017)Bahat, Efrat, and Irani]{bahat2017non}
Yuval Bahat, Netalee Efrat, and Michal Irani.
\newblock Non-uniform blind deblurring by reblurring.
\newblock In \emph{Proceedings of the IEEE international conference on computer vision}, pp.\  3286--3294, 2017.

\bibitem[Bi{\'n}kowski et~al.(2018)Bi{\'n}kowski, Sutherland, Arbel, and Gretton]{binkowski2018demystifying}
Miko{\l}aj Bi{\'n}kowski, Danica~J Sutherland, Michael Arbel, and Arthur Gretton.
\newblock Demystifying mmd gans.
\newblock \emph{arXiv preprint arXiv:1801.01401}, 2018.

\bibitem[Blau et~al.(2018)Blau, Mechrez, Timofte, Michaeli, and Zelnik-Manor]{blau20182018}
Yochai Blau, Roey Mechrez, Radu Timofte, Tomer Michaeli, and Lihi Zelnik-Manor.
\newblock The 2018 pirm challenge on perceptual image super-resolution.
\newblock In \emph{Proceedings of the European conference on computer vision (ECCV) workshops}, pp.\  0--0, 2018.

\bibitem[Chen et~al.(2022)Chen, Chu, Zhang, and Sun]{chen2022simple}
Liangyu Chen, Xiaojie Chu, Xiangyu Zhang, and Jian Sun.
\newblock Simple baselines for image restoration.
\newblock In \emph{European conference on computer vision}, pp.\  17--33. Springer, 2022.

\bibitem[Chen et~al.(2011)Chen, He, Yang, and Wu]{chen2011effective}
Xiaogang Chen, Xiangjian He, Jie Yang, and Qiang Wu.
\newblock An effective document image deblurring algorithm.
\newblock In \emph{CVPR 2011}, pp.\  369--376. IEEE, 2011.

\bibitem[Cho et~al.(2012)Cho, Wang, and Lee]{cho2012text}
Hojin Cho, Jue Wang, and Seungyong Lee.
\newblock Text image deblurring using text-specific properties.
\newblock In \emph{Computer Vision--ECCV 2012: 12th European Conference on Computer Vision, Florence, Italy, October 7-13, 2012, Proceedings, Part V 12}, pp.\  524--537. Springer, 2012.

\bibitem[Cho et~al.(2021)Cho, Ji, Hong, Jung, and Ko]{cho2021rethinking}
Sung-Jin Cho, Seo-Won Ji, Jun-Pyo Hong, Seung-Won Jung, and Sung-Jea Ko.
\newblock Rethinking coarse-to-fine approach in single image deblurring.
\newblock In \emph{Proceedings of the IEEE/CVF international conference on computer vision}, pp.\  4641--4650, 2021.

\bibitem[Dosovitskiy et~al.(2020)Dosovitskiy, Beyer, Kolesnikov, Weissenborn, Zhai, Unterthiner, Dehghani, Minderer, Heigold, Gelly, et~al.]{dosovitskiy2020image}
Alexey Dosovitskiy, Lucas Beyer, Alexander Kolesnikov, Dirk Weissenborn, Xiaohua Zhai, Thomas Unterthiner, Mostafa Dehghani, Matthias Minderer, Georg Heigold, Sylvain Gelly, et~al.
\newblock An image is worth 16x16 words: Transformers for image recognition at scale.
\newblock \emph{arXiv preprint arXiv:2010.11929}, 2020.

\bibitem[Fang et~al.(2023)Fang, Wu, Dong, Li, Wu, and Shi]{fang2023self}
Zhenxuan Fang, Fangfang Wu, Weisheng Dong, Xin Li, Jinjian Wu, and Guangming Shi.
\newblock Self-supervised non-uniform kernel estimation with flow-based motion prior for blind image deblurring.
\newblock In \emph{Proceedings of the IEEE/CVF conference on computer vision and pattern recognition}, pp.\  18105--18114, 2023.

\bibitem[Fu et~al.(2022)Fu, Zheng, Ma, Ye, Yang, and He]{fu2022edge}
Zhichao Fu, Yingbin Zheng, Tianlong Ma, Hao Ye, Jing Yang, and Liang He.
\newblock Edge-aware deep image deblurring.
\newblock \emph{Neurocomputing}, 502:\penalty0 37--47, 2022.

\bibitem[Gao et~al.(2019)Gao, Tao, Shen, and Jia]{gao2019dynamic}
Hongyun Gao, Xin Tao, Xiaoyong Shen, and Jiaya Jia.
\newblock Dynamic scene deblurring with parameter selective sharing and nested skip connections.
\newblock In \emph{Proceedings of the IEEE/CVF conference on computer vision and pattern recognition}, pp.\  3848--3856, 2019.

\bibitem[Ghildyal \& Liu(2022)Ghildyal and Liu]{ghildyal2022stlpips}
Abhijay Ghildyal and Feng Liu.
\newblock Shift-tolerant perceptual similarity metric.
\newblock In \emph{European Conference on Computer Vision}, pp.\  91--107. Springer, 2022.

\bibitem[Hassani \& Shi(2022)Hassani and Shi]{hassani2022dilated}
Ali Hassani and Humphrey Shi.
\newblock Dilated neighborhood attention transformer.
\newblock \emph{arXiv preprint arXiv:2209.15001}, 2022.

\bibitem[Hassani et~al.(2023)Hassani, Walton, Li, Li, and Shi]{hassani2023neighborhood}
Ali Hassani, Steven Walton, Jiachen Li, Shen Li, and Humphrey Shi.
\newblock Neighborhood attention transformer.
\newblock In \emph{Proceedings of the IEEE/CVF Conference on Computer Vision and Pattern Recognition}, pp.\  6185--6194, 2023.

\bibitem[Hendrycks \& Gimpel(2016)Hendrycks and Gimpel]{hendrycks2016gaussian}
Dan Hendrycks and Kevin Gimpel.
\newblock Gaussian error linear units (gelus).
\newblock \emph{arXiv preprint arXiv:1606.08415}, 2016.

\bibitem[Heusel et~al.(2017)Heusel, Ramsauer, Unterthiner, Nessler, and Hochreiter]{heusel2017gans}
Martin Heusel, Hubert Ramsauer, Thomas Unterthiner, Bernhard Nessler, and Sepp Hochreiter.
\newblock Gans trained by a two time-scale update rule converge to a local nash equilibrium.
\newblock \emph{Advances in neural information processing systems}, 30, 2017.

\bibitem[Hirsch et~al.(2011)Hirsch, Schuler, Harmeling, and Sch{\"o}lkopf]{hirsch2011fast}
Michael Hirsch, Christian~J Schuler, Stefan Harmeling, and Bernhard Sch{\"o}lkopf.
\newblock Fast removal of non-uniform camera shake.
\newblock In \emph{2011 International Conference on Computer Vision}, pp.\  463--470. IEEE, 2011.

\bibitem[Hradi{\v{s}} et~al.(2015)Hradi{\v{s}}, Kotera, Zemc{\i}k, and {\v{S}}roubek]{hradivs2015convolutional}
Michal Hradi{\v{s}}, Jan Kotera, Pavel Zemc{\i}k, and Filip {\v{S}}roubek.
\newblock Convolutional neural networks for direct text deblurring.
\newblock In \emph{Proceedings of BMVC}, 2015.

\bibitem[Hu et~al.(2018)Hu, Shen, and Sun]{hu2018squeeze}
Jie Hu, Li~Shen, and Gang Sun.
\newblock Squeeze-and-excitation networks.
\newblock In \emph{Proceedings of the IEEE conference on computer vision and pattern recognition}, pp.\  7132--7141, 2018.

\bibitem[Ji et~al.(2022)Ji, Lee, Kim, Hong, Baek, Jung, and Ko]{ji2022xydeblur}
Seo-Won Ji, Jeongmin Lee, Seung-Wook Kim, Jun-Pyo Hong, Seung-Jin Baek, Seung-Won Jung, and Sung-Jea Ko.
\newblock Xydeblur: divide and conquer for single image deblurring.
\newblock In \emph{Proceedings of the IEEE/CVF conference on computer vision and pattern recognition}, pp.\  17421--17430, 2022.

\bibitem[Jiang et~al.(2021)Jiang, Chang, and Wang]{jiang2021transgan}
Yifan Jiang, Shiyu Chang, and Zhangyang Wang.
\newblock Transgan: Two transformers can make one strong gan.
\newblock \emph{arXiv preprint arXiv:2102.07074}, 1\penalty0 (3), 2021.

\bibitem[Kong et~al.(2023)Kong, Dong, Ge, Li, and Pan]{kong2023efficient}
Lingshun Kong, Jiangxin Dong, Jianjun Ge, Mingqiang Li, and Jinshan Pan.
\newblock Efficient frequency domain-based transformers for high-quality image deblurring.
\newblock In \emph{Proceedings of the IEEE/CVF Conference on Computer Vision and Pattern Recognition}, pp.\  5886--5895, 2023.

\bibitem[Kupyn et~al.(2018)Kupyn, Budzan, Mykhailych, Mishkin, and Matas]{kupyn2018deblurgan}
Orest Kupyn, Volodymyr Budzan, Mykola Mykhailych, Dmytro Mishkin, and Ji{\v{r}}{\'\i} Matas.
\newblock Deblurgan: Blind motion deblurring using conditional adversarial networks.
\newblock In \emph{Proceedings of the IEEE conference on computer vision and pattern recognition}, pp.\  8183--8192, 2018.

\bibitem[Kupyn et~al.(2019)Kupyn, Martyniuk, Wu, and Wang]{kupyn2019deblurgan}
Orest Kupyn, Tetiana Martyniuk, Junru Wu, and Zhangyang Wang.
\newblock Deblurgan-v2: Deblurring (orders-of-magnitude) faster and better.
\newblock In \emph{Proceedings of the IEEE/CVF international conference on computer vision}, pp.\  8878--8887, 2019.

\bibitem[Liu et~al.(2021)Liu, Lin, Cao, Hu, Wei, Zhang, Lin, and Guo]{liu2021swin}
Ze~Liu, Yutong Lin, Yue Cao, Han Hu, Yixuan Wei, Zheng Zhang, Stephen Lin, and Baining Guo.
\newblock Swin transformer: Hierarchical vision transformer using shifted windows.
\newblock In \emph{Proceedings of the IEEE/CVF international conference on computer vision}, pp.\  10012--10022, 2021.

\bibitem[Mao et~al.(2024)Mao, Wang, Xie, Li, and Wang]{mao2024loformer}
Xintian Mao, Jiansheng Wang, Xingran Xie, Qingli Li, and Yan Wang.
\newblock Loformer: Local frequency transformer for image deblurring.
\newblock In \emph{Proceedings of the 32nd ACM International Conference on Multimedia}, pp.\  10382--10391, 2024.

\bibitem[Mittal et~al.(2012)Mittal, Soundararajan, and Bovik]{mittal2012making}
Anish Mittal, Rajiv Soundararajan, and Alan~C Bovik.
\newblock Making a “completely blind” image quality analyzer.
\newblock \emph{IEEE Signal processing letters}, 20\penalty0 (3):\penalty0 209--212, 2012.

\bibitem[Nah et~al.(2017)Nah, Hyun~Kim, and Mu~Lee]{nah2017deep}
Seungjun Nah, Tae Hyun~Kim, and Kyoung Mu~Lee.
\newblock Deep multi-scale convolutional neural network for dynamic scene deblurring.
\newblock In \emph{Proceedings of the IEEE conference on computer vision and pattern recognition}, pp.\  3883--3891, 2017.

\bibitem[Purohit \& Rajagopalan(2020)Purohit and Rajagopalan]{purohit2020region}
Kuldeep Purohit and AN~Rajagopalan.
\newblock Region-adaptive dense network for efficient motion deblurring.
\newblock In \emph{Proceedings of the AAAI conference on artificial intelligence}, pp.\  11882--11889, 2020.

\bibitem[Rim et~al.(2020)Rim, Lee, Won, and Cho]{rim2020real}
Jaesung Rim, Haeyun Lee, Jucheol Won, and Sunghyun Cho.
\newblock Real-world blur dataset for learning and benchmarking deblurring algorithms.
\newblock In \emph{Computer Vision--ECCV 2020: 16th European Conference, Glasgow, UK, August 23--28, 2020, Proceedings, Part XXV 16}, pp.\  184--201. Springer, 2020.

\bibitem[Shen et~al.(2019)Shen, Wang, Lu, Shen, Ling, Xu, and Shao]{shen2019human}
Ziyi Shen, Wenguan Wang, Xiankai Lu, Jianbing Shen, Haibin Ling, Tingfa Xu, and Ling Shao.
\newblock Human-aware motion deblurring.
\newblock In \emph{Proceedings of the IEEE/CVF international conference on computer vision}, pp.\  5572--5581, 2019.

\bibitem[Svoboda et~al.(2016)Svoboda, Hradi{\v{s}}, Mar{\v{s}}{\'\i}k, and Zemc{\'\i}k]{svoboda2016cnn}
Pavel Svoboda, Michal Hradi{\v{s}}, Luk{\'a}{\v{s}} Mar{\v{s}}{\'\i}k, and Pavel Zemc{\'\i}k.
\newblock Cnn for license plate motion deblurring.
\newblock In \emph{2016 IEEE International Conference on Image Processing (ICIP)}, pp.\  3832--3836. IEEE, 2016.

\bibitem[Tao et~al.(2018)Tao, Gao, Shen, Wang, and Jia]{tao2018scale}
Xin Tao, Hongyun Gao, Xiaoyong Shen, Jue Wang, and Jiaya Jia.
\newblock Scale-recurrent network for deep image deblurring.
\newblock In \emph{Proceedings of the IEEE conference on computer vision and pattern recognition}, pp.\  8174--8182, 2018.

\bibitem[Tsai et~al.(2022)Tsai, Peng, Lin, Tsai, and Lin]{tsai2022stripformer}
Fu-Jen Tsai, Yan-Tsung Peng, Yen-Yu Lin, Chung-Chi Tsai, and Chia-Wen Lin.
\newblock Stripformer: Strip transformer for fast image deblurring.
\newblock In \emph{European conference on computer vision}, pp.\  146--162. Springer, 2022.

\bibitem[Vaswani(2017)]{vaswani2017attention}
A~Vaswani.
\newblock Attention is all you need.
\newblock \emph{Advances in Neural Information Processing Systems}, 2017.

\bibitem[Wang et~al.(2020)Wang, Wu, Zhu, Li, Zuo, and Hu]{wang2020eca}
Qilong Wang, Banggu Wu, Pengfei Zhu, Peihua Li, Wangmeng Zuo, and Qinghua Hu.
\newblock Eca-net: Efficient channel attention for deep convolutional neural networks.
\newblock In \emph{Proceedings of the IEEE/CVF conference on computer vision and pattern recognition}, pp.\  11534--11542, 2020.

\bibitem[Wang et~al.(2022)Wang, Cun, Bao, Zhou, Liu, and Li]{wang2022uformer}
Zhendong Wang, Xiaodong Cun, Jianmin Bao, Wengang Zhou, Jianzhuang Liu, and Houqiang Li.
\newblock Uformer: A general u-shaped transformer for image restoration.
\newblock In \emph{Proceedings of the IEEE/CVF conference on computer vision and pattern recognition}, pp.\  17683--17693, 2022.

\bibitem[Whang et~al.(2022)Whang, Delbracio, Talebi, Saharia, Dimakis, and Milanfar]{whang2022deblurring}
Jay Whang, Mauricio Delbracio, Hossein Talebi, Chitwan Saharia, Alexandros~G Dimakis, and Peyman Milanfar.
\newblock Deblurring via stochastic refinement.
\newblock In \emph{Proceedings of the IEEE/CVF Conference on Computer Vision and Pattern Recognition}, pp.\  16293--16303, 2022.

\bibitem[Xu \& Jia(2010)Xu and Jia]{xu2010two}
Li~Xu and Jiaya Jia.
\newblock Two-phase kernel estimation for robust motion deblurring.
\newblock In \emph{Computer Vision--ECCV 2010: 11th European Conference on Computer Vision, Heraklion, Crete, Greece, September 5-11, 2010, Proceedings, Part I 11}, pp.\  157--170. Springer, 2010.

\bibitem[Yan et~al.(2017)Yan, Ren, Guo, Wang, and Cao]{yan2017image}
Yanyang Yan, Wenqi Ren, Yuanfang Guo, Rui Wang, and Xiaochun Cao.
\newblock Image deblurring via extreme channels prior.
\newblock In \emph{Proceedings of the IEEE conference on computer vision and pattern recognition}, pp.\  4003--4011, 2017.

\bibitem[Yuan et~al.(2021)Yuan, Chen, Wang, Yu, Shi, Jiang, Tay, Feng, and Yan]{yuan2021tokens}
Li~Yuan, Yunpeng Chen, Tao Wang, Weihao Yu, Yujun Shi, Zi-Hang Jiang, Francis~EH Tay, Jiashi Feng, and Shuicheng Yan.
\newblock Tokens-to-token vit: Training vision transformers from scratch on imagenet.
\newblock In \emph{Proceedings of the IEEE/CVF international conference on computer vision}, pp.\  558--567, 2021.

\bibitem[Zamir et~al.(2021)Zamir, Arora, Khan, Hayat, Khan, Yang, and Shao]{zamir2021multi}
Syed~Waqas Zamir, Aditya Arora, Salman Khan, Munawar Hayat, Fahad~Shahbaz Khan, Ming-Hsuan Yang, and Ling Shao.
\newblock Multi-stage progressive image restoration.
\newblock In \emph{Proceedings of the IEEE/CVF conference on computer vision and pattern recognition}, pp.\  14821--14831, 2021.

\bibitem[Zamir et~al.(2022)Zamir, Arora, Khan, Hayat, Khan, and Yang]{zamir2022restormer}
Syed~Waqas Zamir, Aditya Arora, Salman Khan, Munawar Hayat, Fahad~Shahbaz Khan, and Ming-Hsuan Yang.
\newblock Restormer: Efficient transformer for high-resolution image restoration.
\newblock In \emph{Proceedings of the IEEE/CVF conference on computer vision and pattern recognition}, pp.\  5728--5739, 2022.

\bibitem[Zhang et~al.(2019)Zhang, Dai, Li, and Koniusz]{zhang2019deep}
Hongguang Zhang, Yuchao Dai, Hongdong Li, and Piotr Koniusz.
\newblock Deep stacked hierarchical multi-patch network for image deblurring.
\newblock In \emph{Proceedings of the IEEE/CVF conference on computer vision and pattern recognition}, pp.\  5978--5986, 2019.

\bibitem[Zhang et~al.(2020)Zhang, Luo, Zhong, Ma, Stenger, Liu, and Li]{zhang2020deblurring}
Kaihao Zhang, Wenhan Luo, Yiran Zhong, Lin Ma, Bjorn Stenger, Wei Liu, and Hongdong Li.
\newblock Deblurring by realistic blurring.
\newblock In \emph{Proceedings of the IEEE/CVF conference on computer vision and pattern recognition}, pp.\  2737--2746, 2020.

\bibitem[Zhang et~al.(2022)Zhang, Ren, Luo, Lai, Stenger, Yang, and Li]{zhang2022deep}
Kaihao Zhang, Wenqi Ren, Wenhan Luo, Wei-Sheng Lai, Bj{\"o}rn Stenger, Ming-Hsuan Yang, and Hongdong Li.
\newblock Deep image deblurring: A survey.
\newblock \emph{International Journal of Computer Vision}, 130\penalty0 (9):\penalty0 2103--2130, 2022.

\bibitem[Zhang et~al.(2018)Zhang, Isola, Efros, Shechtman, and Wang]{zhang2018unreasonable}
Richard Zhang, Phillip Isola, Alexei~A Efros, Eli Shechtman, and Oliver Wang.
\newblock The unreasonable effectiveness of deep features as a perceptual metric.
\newblock In \emph{Proceedings of the IEEE conference on computer vision and pattern recognition}, pp.\  586--595, 2018.

\bibitem[Zou et~al.(2021)Zou, Jiang, Zhang, Chen, Lu, and Wu]{zou2021sdwnet}
Wenbin Zou, Mingchao Jiang, Yunchen Zhang, Liang Chen, Zhiyong Lu, and Yi~Wu.
\newblock Sdwnet: A straight dilated network with wavelet transformation for image deblurring.
\newblock In \emph{Proceedings of the IEEE/CVF international conference on computer vision}, pp.\  1895--1904, 2021.

\bibitem[Zvezdakova et~al.(2019)Zvezdakova, Kulikov, Kondranin, and Vatolin]{zvezdakova2019barriers}
Anastasia Zvezdakova, Dmitriy Kulikov, Denis Kondranin, and Dmitriy Vatolin.
\newblock Barriers towards no-reference metrics application to compressed video quality analysis: On the example of no-reference metric niqe.
\newblock \emph{arXiv preprint arXiv:1907.03842}, 2019.

\end{thebibliography}
\bibliographystyle{tmlr}


\end{document}


\maketitle
In this document, we firstly provide further analysis on lightweight dual-stage feature fusion (LDFF) and divide-and-multiply feed-forward network (DMFN).
Next, we present more training details and dataset introduction that are not included in the main manuscript. Finally, we test DeblurDiNAT (both the small and large variant) on real-world blurry image dataset RWBI~\citep{zhang2020deblurring},
and show more deblurred images and comparisons on various real-world blurred scenes~\citep{zhang2020deblurring,rim2020real}.

\section{Further Analysis on LDFF}
In the main paper, we have discussed the efficacy and efficiency of LDFF, a novel feature fusion method designed for image restoration tasks. We firsty design LDFF for multi-scale feature fusion and extend it for same-scale feature fusion. We only compare it with Asymmetric Feature Fusion (AFF)~\citep{cho2021rethinking} for multi-scale features and its simplified version for same-scale features, because these approaches are plug-and-play modules that can be directly incorporated into DeblurDiNAT.

\section{Further Analysis on DMFN}
While we remove GeLU from Decoder Gated FFN (GDFN)~\citep{zamir2022restormer},
non-linearity is still preserved in our network 
because feature maps already pass through Encoders with LeakyReLU and Fusion Blocks with GeLU.
GDFN is formulated as, $X_1\odot (X_2\Phi(X_2))$,
where $\Phi$ is standard Gaussian cumulative distribution.
Our method is essentially to remove $\Phi$ from Decoder FFN in a Transformer architecture.

\section{More Implementation Details and Datasets}

\noindent \textbf{Implementation Details.} We propose two variants of our model, DeblurDiNAT-S and DeblurDiNAT-L.
For both of them,
the global dilation factors are $[9, 18, 36]$.
We use the Pytorch DataParallel for multi-GPU training.
The training time on 8 NVIDIA A100 GPUs was around 5 days for DeblurDiNAT-S, 
and 9 days for DeblurDiNAT-L.

\noindent \textbf{GoPro.} The GoPro dataset~\citep{nah2017deep} consists of 3214 image pairs with resolution of $1280\times720$,
where 2103 pairs are for training and the rest are for test.
Instead of using a blur kernel to convolve on a sharp image,
Nah \etal~\citep{nah2017deep} collected sharp images from 240 fps videos
and averaged several numbers of sequential frames to generate blurred images.

\noindent \textbf{HIDE.} The HIDE dataset~\citep{shen2019human} covers both wide-range and close-range scenes, and is especially for human-aware (pedestrian) motion deblurring.
Its test set contains 2025 image pairs.
The blurred images were generated by averaging 11 successive frames from videos,
and its synthesis method is similar to the one used by GoPro~\citep{nah2017deep}.

\noindent \textbf{RealBlur.} RealBlur~\citep{rim2020real} is a large-scale real-world deblurring dataset,
consisting of two subsets, RealBlur-J and RealBlur-R.
Each subset provides 3758 image pairs for training and 980 pairs for test.
\cite{rim2020real} captured real-world images blurred by camera shakes
in low-light environments.
The paired images in RealBlur datasets are not well aligned.

\noindent \textbf{RWBI.} The RWBI dataset~\citep{zhang2020deblurring} contains 3112 diverse blurred images,
and 1000 of them are for test.
Zhang \etal\citep{zhang2020deblurring} captured blurry images with different hand-held devices,
and there is no ground truth within the dataset. 

\section{Additional Visual Results}
\begin{figure*}[tb]
  \captionsetup[subfigure]{labelformat=empty, justification=centering}
  \centering
  \begin{subfigure}[b]{0.3\textwidth}
    \centering
    \includegraphics[width=\textwidth]{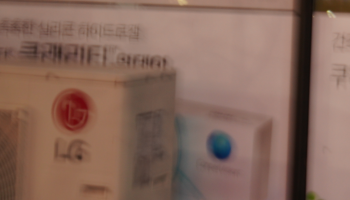}
    \caption{Blurred}
  \end{subfigure}
  \hfill
  \begin{subfigure}[b]{0.3\textwidth}
    \centering
    \includegraphics[width=\textwidth]{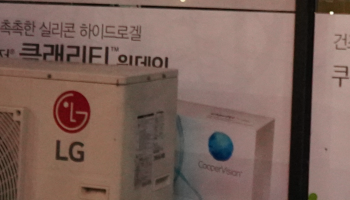}
    \caption{Ground Truth}
  \end{subfigure}
  \hfill
  \begin{subfigure}[b]{0.3\textwidth}
      \centering
      \includegraphics[width=\textwidth]{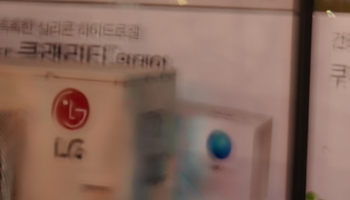}
      \caption{MIMO-Unet+~\citeyear{cho2021rethinking}}
  \end{subfigure}
  \hfill
  \begin{subfigure}[b]{0.3\textwidth}
      \centering
      \includegraphics[width=\textwidth]{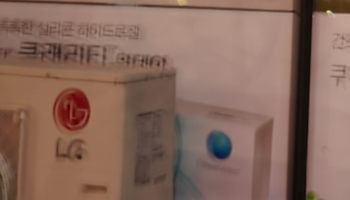}
      \caption{NAFNet~\citeyear{chen2022simple}}
  \end{subfigure}
  \hfill
  \begin{subfigure}[b]{0.3\textwidth}
      \centering
      \includegraphics[width=\textwidth]{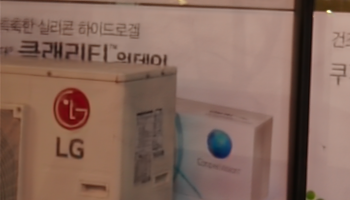}
      \caption{Uformer-B~\citeyear{wang2022uformer}}
  \end{subfigure}
  \hfill
  \begin{subfigure}[b]{0.3\textwidth}
      \centering
      \includegraphics[width=\textwidth]{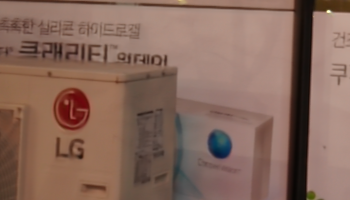}
      \caption{Restormer~\citeyear{zamir2022restormer}}
  \end{subfigure}
  \hfill
  \begin{subfigure}[b]{0.3\textwidth}
      \centering
      \includegraphics[width=\textwidth]{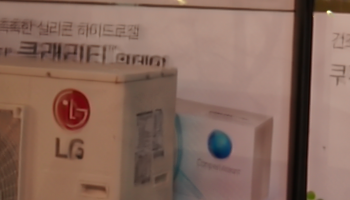}
      \caption{Stripformer~\citeyear{tsai2022stripformer}}
  \end{subfigure}
  \hfill
  \begin{subfigure}[b]{0.3\textwidth}
      \centering
      \includegraphics[width=\textwidth]{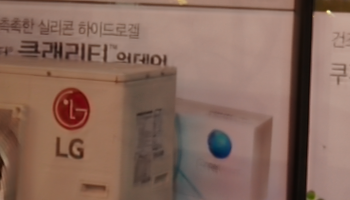}
      \caption{FFTformer~\citeyear{kong2023efficient}}
  \end{subfigure}
  \hfill
  \begin{subfigure}[b]{0.3\textwidth}
      \centering
      \includegraphics[width=\textwidth]{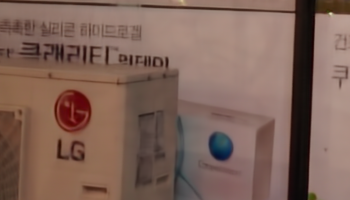}
      \caption{UFPNet~\citeyear{fang2023self}}
  \end{subfigure}
  \hfill
  \begin{subfigure}[b]{0.3\textwidth}
      \centering
      \includegraphics[width=\textwidth]{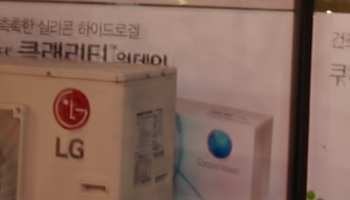}
      \caption{LoFormer~\citeyear{mao2024loformer}}
  \end{subfigure}
  \hfill
  \begin{subfigure}[b]{0.3\textwidth}
      \centering
      \includegraphics[width=\textwidth]{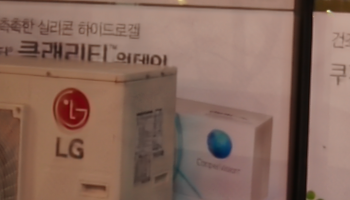}
      \caption{\bftab DeblurDiNAT-S}
  \end{subfigure}
  \hfill
  \begin{subfigure}[b]{0.3\textwidth}
      \centering
      \includegraphics[width=\textwidth]{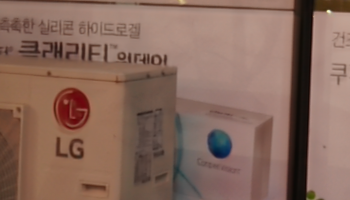}
      \caption{\bftab DeblurDiNAT-L}
  \end{subfigure}
     \caption{Visual comparisons on the RealBlur-J dataset~\citep{rim2020real}. LoFormer~\citep{mao2024loformer} with 49M parameters and DeblurDiNAT-L with 16.1M parameters generate the best image quality.}\label{fig:realj1}
\end{figure*}
In this section,
we provide more visual comparisons on RWBI~\citep{zhang2020deblurring} 
and RealBlur datasets~\citep{rim2020real}. GoPro-trained networks are directly applied to these datasets.

\begin{figure*}[tb]
  \captionsetup[subfigure]{labelformat=empty, justification=centering}
  \centering
  \begin{subfigure}[b]{0.45\textwidth}
    \centering
    \includegraphics[width=\textwidth]{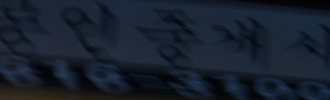}
    \caption{Blurred}
  \end{subfigure}
  \hfill
  \begin{subfigure}[b]{0.45\textwidth}
    \centering
    \includegraphics[width=\textwidth]{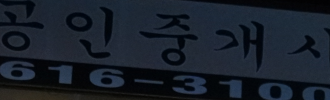}
    \caption{Ground Truth}
  \end{subfigure}
  \hfill
  \begin{subfigure}[b]{0.45\textwidth}
      \centering
      \includegraphics[width=\textwidth]{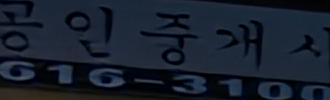}
      \caption{DeblurGANv2~\citep{kupyn2019deblurgan}}
  \end{subfigure}
  \hfill
  \begin{subfigure}[b]{0.45\textwidth}
      \centering
      \includegraphics[width=\textwidth]{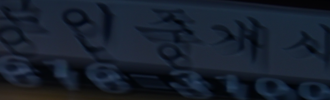}
      \caption{MIMO-Unet+~\citep{cho2021rethinking}}
  \end{subfigure}
  \hfill
  \begin{subfigure}[b]{0.45\textwidth}
      \centering
      \includegraphics[width=\textwidth]{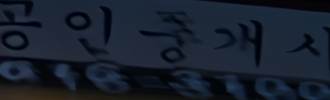}
      \caption{MPRNet~\citep{zamir2021multi}}
  \end{subfigure}
  \hfill
  \begin{subfigure}[b]{0.45\textwidth}
      \centering
      \includegraphics[width=\textwidth]{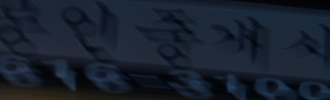}
      \caption{NAFNet~\citep{chen2022simple}}
  \end{subfigure}
  \hfill
  \begin{subfigure}[b]{0.45\textwidth}
      \centering
      \includegraphics[width=\textwidth]{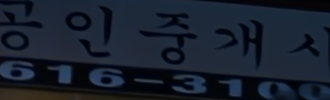}
      \caption{Uformer-B~\citep{wang2022uformer}}
  \end{subfigure}
  \hfill
  \begin{subfigure}[b]{0.45\textwidth}
      \centering
      \includegraphics[width=\textwidth]{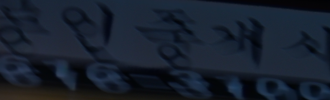}
      \caption{Restormer~\citep{zamir2022restormer}}
  \end{subfigure}
  \hfill
  \begin{subfigure}[b]{0.45\textwidth}
      \centering
      \includegraphics[width=\textwidth]{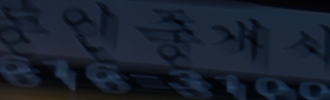}
      \caption{Stripformer~\citep{tsai2022stripformer}}
  \end{subfigure}
  \hfill
  \begin{subfigure}[b]{0.45\textwidth}
      \centering
      \includegraphics[width=\textwidth]{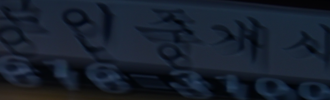}
      \caption{FFTformer~\citep{kong2023efficient}}
  \end{subfigure}
  \hfill
  \begin{subfigure}[b]{0.45\textwidth}
      \centering
      \includegraphics[width=\textwidth]{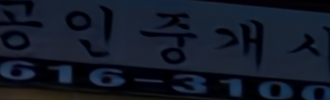}
      \caption{UFPNet~\citep{fang2023self}}
  \end{subfigure}
  \hfill
  \begin{subfigure}[b]{0.45\textwidth}
      \centering
      \includegraphics[width=\textwidth]{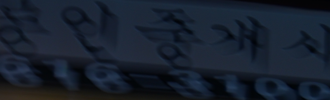}
      \caption{LoFormer~\citep{mao2024loformer}}
  \end{subfigure}
  \hfill
  \begin{subfigure}[b]{0.45\textwidth}
      \centering
      \includegraphics[width=\textwidth]{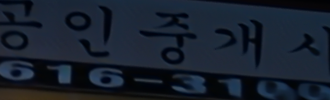}
      \caption{\bftab DeblurDiNAT-S}
  \end{subfigure}
  \hfill
  \begin{subfigure}[b]{0.45\textwidth}
      \centering
      \includegraphics[width=\textwidth]{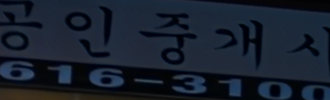}
      \caption{\bftab DeblurDiNAT-L}
  \end{subfigure}
     \caption{Qualitative comparisons on the RealBlur-J dataset~\citep{rim2020real}.}\label{fig:realj2}
\end{figure*}

\begin{figure*}[tb]
  \captionsetup[subfigure]{labelformat=empty, justification=centering}
  \centering
  \begin{subfigure}[b]{0.3\textwidth}
    \centering
    \includegraphics[width=\textwidth]{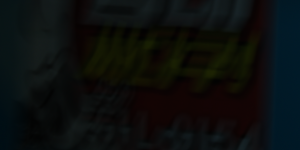}
    \caption{Blurred}
  \end{subfigure}
  \hfill
  \begin{subfigure}[b]{0.3\textwidth}
    \centering
    \includegraphics[width=\textwidth]{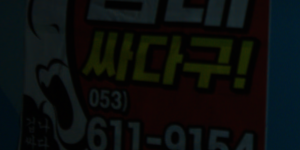}
    \caption{Ground Truth}
  \end{subfigure}
  \hfill
  \begin{subfigure}[b]{0.3\textwidth}
      \centering
      \includegraphics[width=\textwidth]{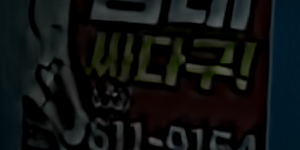}
      \caption{DeblurGANv2~\citeyear{kupyn2019deblurgan}}
  \end{subfigure}
  \hfill
  \begin{subfigure}[b]{0.3\textwidth}
      \centering
      \includegraphics[width=\textwidth]{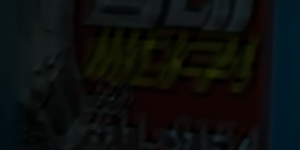}
      \caption{MIMO-Unet+~\citeyear{cho2021rethinking}}
  \end{subfigure}
  \hfill
  \begin{subfigure}[b]{0.3\textwidth}
      \centering
      \includegraphics[width=\textwidth]{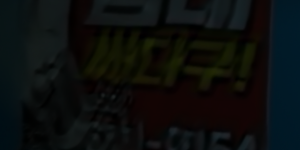}
      \caption{MPRNet~\citeyear{zamir2021multi}}
  \end{subfigure}
  \hfill
  \begin{subfigure}[b]{0.3\textwidth}
      \centering
      \includegraphics[width=\textwidth]{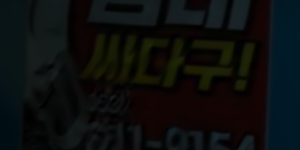}
      \caption{NAFNet~\citeyear{chen2022simple}}
  \end{subfigure}
  \hfill
  \begin{subfigure}[b]{0.3\textwidth}
      \centering
      \includegraphics[width=\textwidth]{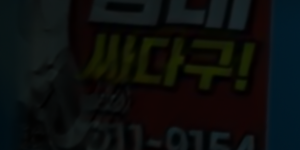}
      \caption{Uformer-B~\citeyear{wang2022uformer}}
  \end{subfigure}
  \hfill
  \begin{subfigure}[b]{0.3\textwidth}
      \centering
      \includegraphics[width=\textwidth]{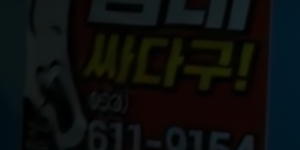}
      \caption{Restormer~\citeyear{zamir2022restormer}}
  \end{subfigure}
  \hfill
  \begin{subfigure}[b]{0.3\textwidth}
      \centering
      \includegraphics[width=\textwidth]{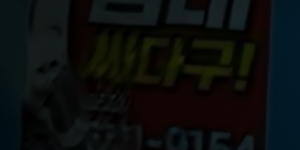}
      \caption{Stripformer~\citeyear{tsai2022stripformer}}
  \end{subfigure}
  \hfill
  \begin{subfigure}[b]{0.3\textwidth}
      \centering
      \includegraphics[width=\textwidth]{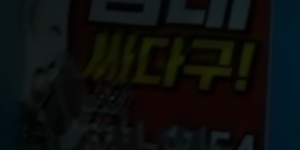}
      \caption{FFTformer~\citeyear{kong2023efficient}}
  \end{subfigure}
  \hfill
  \begin{subfigure}[b]{0.3\textwidth}
      \centering
      \includegraphics[width=\textwidth]{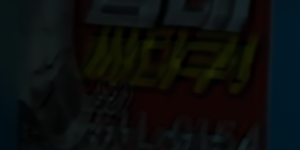}
      \caption{UFPNet~\citeyear{fang2023self}}
  \end{subfigure}
  \hfill
  \begin{subfigure}[b]{0.3\textwidth}
      \centering
      \includegraphics[width=\textwidth]{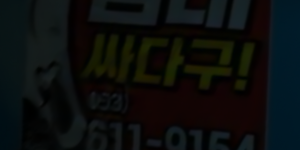}
      \caption{LoFormer~\citeyear{mao2024loformer}}
  \end{subfigure}
  \hfill
  \begin{subfigure}[b]{0.3\textwidth}
      \centering
      \includegraphics[width=\textwidth]{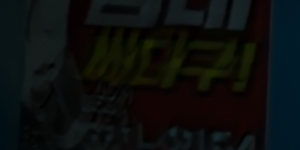}
      \caption{\bftab DeblurDiNAT-S}
  \end{subfigure}
  \hfill
  \begin{subfigure}[b]{0.3\textwidth}
      \centering
      \includegraphics[width=\textwidth]{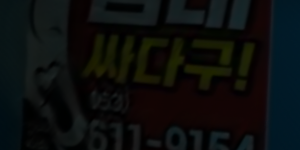}
      \caption{\bftab DeblurDiNAT-L}
  \end{subfigure}
     \caption{Qualitative comparisons on the RealBlur-R dataset~\citep{rim2020real}.}\label{fig:realr1}
\end{figure*}

\begin{figure*}[tb]
  \captionsetup[subfigure]{labelformat=empty, justification=centering}
  \centering
  \begin{subfigure}[b]{0.3\textwidth}
    \centering
    \includegraphics[width=\textwidth]{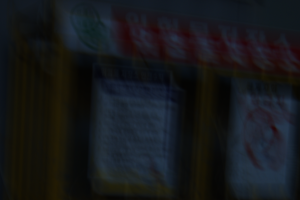}
    \caption{Blurred}
  \end{subfigure}
  \hfill
  \begin{subfigure}[b]{0.3\textwidth}
    \centering
    \includegraphics[width=\textwidth]{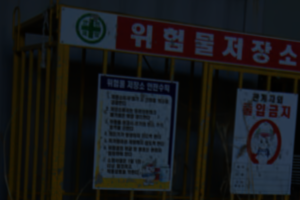}
    \caption{Ground Truth}
  \end{subfigure}
  \hfill
  \begin{subfigure}[b]{0.3\textwidth}
      \centering
      \includegraphics[width=\textwidth]{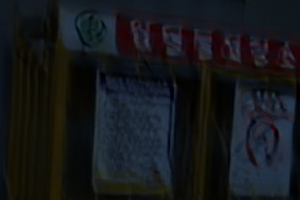}
      \caption{DeblurGANv2~\citeyear{kupyn2019deblurgan}}
  \end{subfigure}
  \hfill
  \begin{subfigure}[b]{0.3\textwidth}
      \centering
      \includegraphics[width=\textwidth]{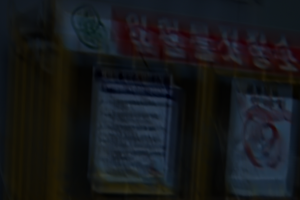}
      \caption{MIMO-Unet+~\citeyear{cho2021rethinking}}
  \end{subfigure}
  \hfill
  \begin{subfigure}[b]{0.3\textwidth}
      \centering
      \includegraphics[width=\textwidth]{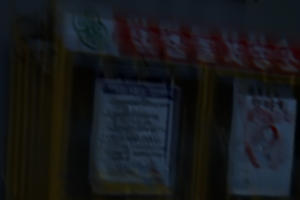}
      \caption{MPRNet~\citeyear{zamir2021multi}}
  \end{subfigure}
  \hfill
  \begin{subfigure}[b]{0.3\textwidth}
      \centering
      \includegraphics[width=\textwidth]{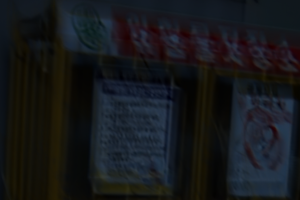}
      \caption{NAFNet~\citeyear{chen2022simple}}
  \end{subfigure}
  \hfill
  \begin{subfigure}[b]{0.3\textwidth}
      \centering
      \includegraphics[width=\textwidth]{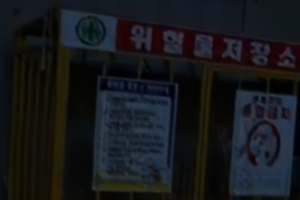}
      \caption{Uformer-B~\citeyear{wang2022uformer}}
  \end{subfigure}
  \hfill
  \begin{subfigure}[b]{0.3\textwidth}
      \centering
      \includegraphics[width=\textwidth]{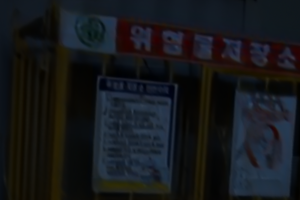}
      \caption{Restormer~\citeyear{zamir2022restormer}}
  \end{subfigure}
  \hfill
  \begin{subfigure}[b]{0.3\textwidth}
      \centering
      \includegraphics[width=\textwidth]{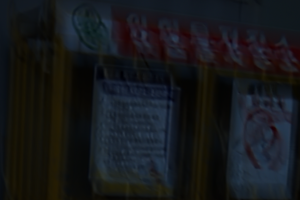}
      \caption{Stripformer~\citeyear{tsai2022stripformer}}
  \end{subfigure}
  \hfill
  \begin{subfigure}[b]{0.3\textwidth}
      \centering
      \includegraphics[width=\textwidth]{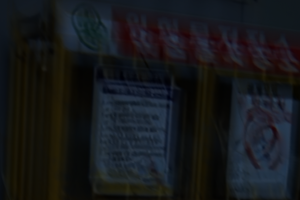}
      \caption{FFTformer~\citeyear{kong2023efficient}}
  \end{subfigure}
  \hfill
  \begin{subfigure}[b]{0.3\textwidth}
      \centering
      \includegraphics[width=\textwidth]{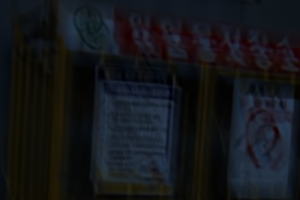}
      \caption{UFPNet~\citeyear{fang2023self}}
  \end{subfigure}
  \hfill
  \begin{subfigure}[b]{0.3\textwidth}
      \centering
      \includegraphics[width=\textwidth]{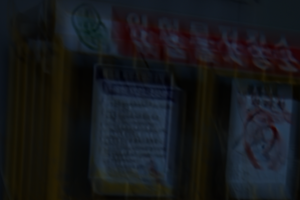}
      \caption{LoFormer~\citeyear{mao2024loformer}}
  \end{subfigure}
  \hfill
  \begin{subfigure}[b]{0.3\textwidth}
      \centering
      \includegraphics[width=\textwidth]{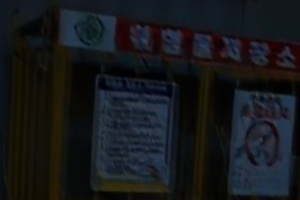}
      \caption{\bftab DeblurDiNAT-S}
  \end{subfigure}
  \hfill
  \begin{subfigure}[b]{0.3\textwidth}
      \centering
      \includegraphics[width=\textwidth]{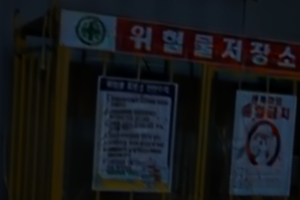}
      \caption{\bftab DeblurDiNAT-L}
  \end{subfigure}
     \caption{Qualitative comparisons on the RealBlur-R dataset~\citep{rim2020real}.}\label{fig:realr2}
\end{figure*}

\begin{figure*}[tb]
  \captionsetup[subfigure]{labelformat=empty, justification=centering}
  \centering
  \begin{subfigure}[b]{0.3\textwidth}
    \centering
    \includegraphics[width=\textwidth]{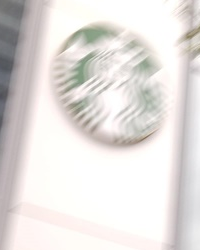}
    \caption{Blurred}
  \end{subfigure}
  \hfill
  \begin{subfigure}[b]{0.3\textwidth}
      \centering
      \includegraphics[width=\textwidth]{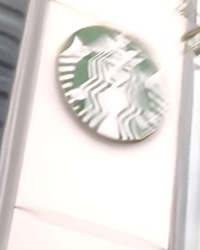}
      \caption{NAFNet~\citeyear{chen2022simple}}
  \end{subfigure}
  \hfill
  \begin{subfigure}[b]{0.3\textwidth}
      \centering
      \includegraphics[width=\textwidth]{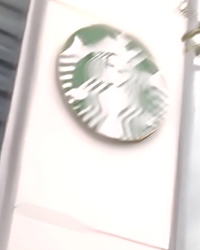}
      \caption{Uformer-B~\citeyear{wang2022uformer}}
  \end{subfigure}
  \hfill
  \begin{subfigure}[b]{0.3\textwidth}
      \centering
      \includegraphics[width=\textwidth]{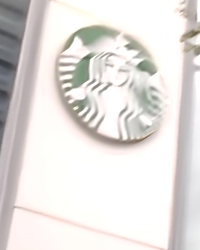}
      \caption{Restormer~\citeyear{zamir2022restormer}}
  \end{subfigure}
  \hfill
  \begin{subfigure}[b]{0.3\textwidth}
      \centering
      \includegraphics[width=\textwidth]{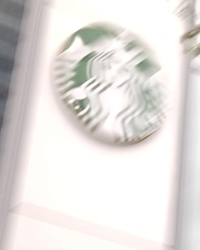}
      \caption{FFTformer~\citeyear{kong2023efficient}}
  \end{subfigure}
  \hfill
  \begin{subfigure}[b]{0.3\textwidth}
      \centering
      \includegraphics[width=\textwidth]{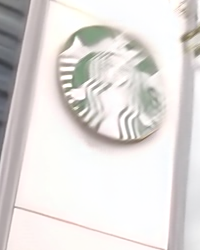}
      \caption{UFPNet~\citeyear{fang2023self}}
  \end{subfigure}
  \hfill
  \begin{subfigure}[b]{0.3\textwidth}
      \centering
      \includegraphics[width=\textwidth]{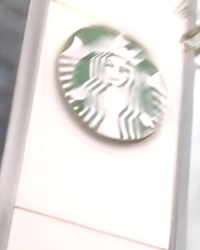}
      \caption{LoFormer~\citeyear{mao2024loformer}}
  \end{subfigure}
  \hfill
  \begin{subfigure}[b]{0.3\textwidth}
      \centering
      \includegraphics[width=\textwidth]{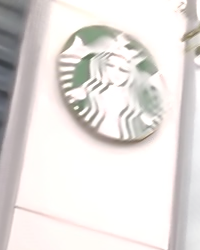}
      \caption{\bftab DeblurDiNAT-S}
  \end{subfigure}
  \hfill
  \begin{subfigure}[b]{0.3\textwidth}
      \centering
      \includegraphics[width=\textwidth]{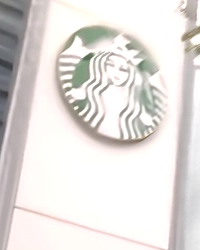}
      \caption{\bftab DeblurDiNAT-L}
  \end{subfigure}
     \caption{Qualitative comparisons on the RWBI dataset~\citep{zhang2020deblurring}.}\label{fig:rwbi1}
\end{figure*}

\begin{figure*}[tb]
  \captionsetup[subfigure]{labelformat=empty, justification=centering}
  \centering
  \begin{subfigure}[b]{0.26\textwidth}
    \centering
    \includegraphics[width=\textwidth]{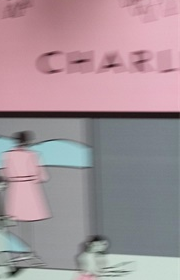}
    \caption{Blurred}
  \end{subfigure}
  \hfill
  \begin{subfigure}[b]{0.26\textwidth}
      \centering
      \includegraphics[width=\textwidth]{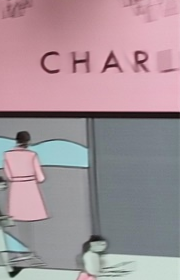}
      \caption{NAFNet~\citeyear{chen2022simple}}
  \end{subfigure}
  \hfill
  \begin{subfigure}[b]{0.26\textwidth}
      \centering
      \includegraphics[width=\textwidth]{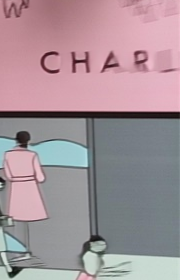}
      \caption{Uformer-B~\citeyear{wang2022uformer}}
  \end{subfigure}
  \hfill
  \begin{subfigure}[b]{0.26\textwidth}
      \centering
      \includegraphics[width=\textwidth]{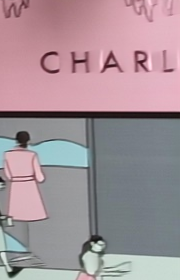}
      \caption{Restormer~\citeyear{zamir2022restormer}}
  \end{subfigure}
  \hfill
  \begin{subfigure}[b]{0.26\textwidth}
      \centering
      \includegraphics[width=\textwidth]{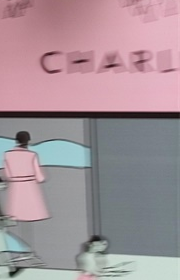}
      \caption{FFTformer~\citeyear{kong2023efficient}}
  \end{subfigure}
  \hfill
  \begin{subfigure}[b]{0.26\textwidth}
      \centering
      \includegraphics[width=\textwidth]{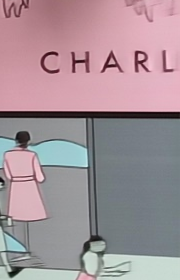}
      \caption{UFPNet~\citeyear{fang2023self}}
  \end{subfigure}
  \hfill
  \begin{subfigure}[b]{0.26\textwidth}
      \centering
      \includegraphics[width=\textwidth]{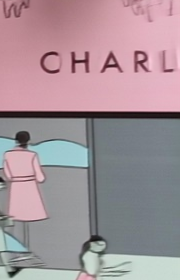}
      \caption{LoFormer~\citeyear{mao2024loformer}}
  \end{subfigure}
  \hfill
  \begin{subfigure}[b]{0.26\textwidth}
      \centering
      \includegraphics[width=\textwidth]{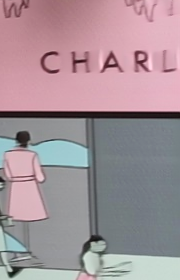}
      \caption{\bftab DeblurDiNAT-S}
  \end{subfigure}
  \hfill
  \begin{subfigure}[b]{0.26\textwidth}
      \centering
      \includegraphics[width=\textwidth]{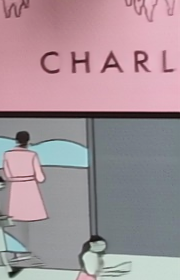}
      \caption{\bftab DeblurDiNAT-L}
  \end{subfigure}
     \caption{Qualitative comparisons on the RWBI dataset~\citep{zhang2020deblurring}.}\label{fig:rwbi2}
\end{figure*}

\begin{figure*}[tb]
  \captionsetup[subfigure]{labelformat=empty, justification=centering}
  \centering
  \begin{subfigure}[b]{0.28\textwidth}
    \centering
    \includegraphics[width=\textwidth]{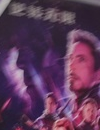}
    \caption{Blurry}
  \end{subfigure}
  \hfill
  \begin{subfigure}[b]{0.26\textwidth}
      \centering
      \includegraphics[width=\textwidth]{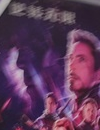}
      \caption{NAFNet~\citeyear{chen2022simple}}
  \end{subfigure}
  \hfill
  \begin{subfigure}[b]{0.28\textwidth}
      \centering
      \includegraphics[width=\textwidth]{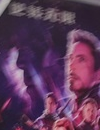}
      \caption{Uformer-B~\citeyear{wang2022uformer}}
  \end{subfigure}
  \hfill
  \begin{subfigure}[b]{0.28\textwidth}
      \centering
      \includegraphics[width=\textwidth]{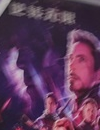}
      \caption{Restormer~\citeyear{zamir2022restormer}}
  \end{subfigure}
  \hfill
  \begin{subfigure}[b]{0.28\textwidth}
      \centering
      \includegraphics[width=\textwidth]{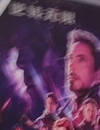}
      \caption{FFTformer~\citeyear{kong2023efficient}}
  \end{subfigure}
  \hfill
  \begin{subfigure}[b]{0.28\textwidth}
      \centering
      \includegraphics[width=\textwidth]{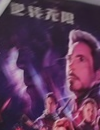}
      \caption{UFPNet~\citeyear{fang2023self}}
  \end{subfigure}
  \hfill
  \begin{subfigure}[b]{0.28\textwidth}
      \centering
      \includegraphics[width=\textwidth]{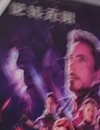}
      \caption{LoFormer~\citeyear{mao2024loformer}}
  \end{subfigure}
  \hfill
  \begin{subfigure}[b]{0.28\textwidth}
      \centering
      \includegraphics[width=\textwidth]{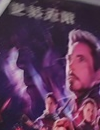}
      \caption{\bftab DeblurDiNAT-S}
  \end{subfigure}
  \hfill
  \begin{subfigure}[b]{0.28\textwidth}
      \centering
      \includegraphics[width=\textwidth]{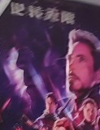}
      \caption{\bftab DeblurDiNAT-L}
  \end{subfigure}
     \caption{Qualitative comparisons on the RWBI dataset~\citep{zhang2020deblurring}.}\label{fig:rwbi4}
\end{figure*}

As shown in Figures~\ref{fig:realj1} - \ref{fig:rwbi4}, 
DeblurDiNAT-L provides sharper edges especially in image regions that have complex patterns.
Specificatlly, Figures~\ref{fig:realj2} - \ref{fig:realr2} show that our model generates deblurred images with the highest visual quality in low-light environments.
As reported in the main manuscript, we notice that UFPNet~\citep{fang2023self} and FFTformer~\citep{kong2023efficient}
introduce unpleasant noises in multiple scenarios,
which have not been observed in the images restored by our network DeblurDiNAT-L.


\clearpage

\clearpage

\clearpage

\clearpage

\clearpage

\clearpage



\bibliography{tmlr}
\bibliographystyle{tmlr}
